\pdfoutput=1
\documentclass[10pt,twocolumn,letterpaper]{article}

\usepackage{arxivstyle}

\usepackage{multirow}
\usepackage{makecell}
\usepackage{array}
\usepackage{adjustbox}
\usepackage{pifont}

\newcommand{\avg}{\textit{Avg.}}
\newcommand{\cmark}{\ding{51}}
\newcommand{\xmark}{\ding{55}}

\renewcommand{\paragraph}[1]{\vspace{.5em}\noindent\textbf{#1.}}

\definecolor{linkblue}{rgb}{0.21,0.49,0.74}
\usepackage[pagebackref,breaklinks,colorlinks,allcolors=linkblue]{hyperref}

\title{FedRAPP: Federated Test-Time Prompt Tuning for Vision-Language Models with Reliability-Aware Prompt Personalization}

\author{Syed Md. Ahnaf Hasan$^{1,*}$, Md Akil Raihan Iftee$^{2}$, Sajib Mistry$^{3}$,\\
Amin Ahsan Ali$^{1}$, AKM Mahbubur Rahman$^{1}$\\[1em]
$^{1}$Center for Computational \& Data Sciences (CCDS), Independent University, Bangladesh\\
$^{2}$Rochester Institute of Technology, USA\\
$^{3}$Curtin University, Australia
}

\begin{document}
\maketitle
{\renewcommand{\thefootnote}{\fnsymbol{footnote}}\footnotetext[1]{Corresponding author: \href{mailto:ahnafhasan335@gmail.com}{ahnafhasan335@gmail.com}}}
\begin{abstract}
Federated prompt learning enables vision-language models like CLIP to collaboratively adapt lightweight soft prompts across distributed clients without sharing raw data. However, existing federated prompt learning methods are designed primarily for supervised settings with labeled client data, and source-free federated prompt tuning at test time remains largely unexplored.
At test time, clients receive unlabeled samples with heterogeneous, often non-IID distributions, causing locally adapted prompts to drift in different directions and reducing the effectiveness of federated aggregation.
In this work, we explore Source-Free Federated Test-Time Prompt Tuning (SF-FTTPT), where distributed clients collaboratively adapt soft prompts on unlabeled test streams without access to source data.
We analyze seven existing federated methods in conjunction with four local test-time prompt tuning methods in this setting, and find that existing federated methods provide limited benefit over independent client adaptation under heterogeneous client distributions, and that the benefit of federation strongly depends on the local adaptation method.
To address this challenge, we propose FedRAPP, a personalized aggregation framework that combines reliability-aware client assessment, server-side exponential moving-average prompt memory, and similarity-guided peer aggregation, so that each client draws on reliable and compatible clients while preserving its local specialization.
Extensive experiments across nine CLIP benchmark datasets and five ImageNet variants demonstrate that FedRAPP consistently outperforms existing federated methods under heterogeneous client distributions, at a small communication and computation overhead.

\end{abstract}

\section{Introduction}
\label{sec}
Vision-language models (VLMs) such as CLIP \cite{radford2021learning} have achieved impressive zero-shot transfer performance by aligning visual and textual representations
in a shared embedding space \cite{jia2021scaling, li2022blip, yuan2021florence}. However, their performance depends critically on the prompts used at inference, and distribution
shifts can substantially degrade their effectiveness. Because end-to-end fine-tuning of these large models is computationally infeasible for edge devices,
parameter-efficient approaches such as prompt tuning optimize task-specific soft prompts while keeping the pretrained VLM
frozen \cite{zhou2022learning, zhou2022conditional, khattak2023maple, jia2022visual}.

\begin{figure}[t]
\centering
\includegraphics[width=0.84\linewidth]{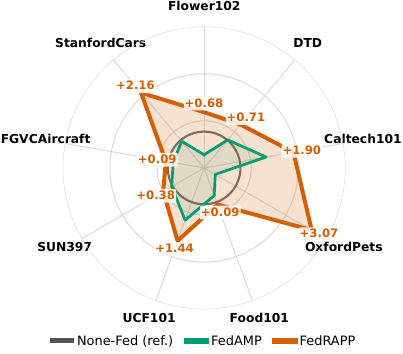}
\vspace{-0.15em}
\caption{Top-1 accuracy gain over None-Fed under the non-IID partition with AdaPrompt local adaptation (Numeric labels show FedRAPP gain over None-Fed).}
\label{fig:intro-radar}
\end{figure}

In practice, the test streams are rarely centralized. Modern AI systems operate across distributed edge environments such as mobile devices, autonomous vehicles,
hospitals, and IoT sensors, where data remains local due to privacy, communication, and latency constraints
\cite{yang2019federated, li2020federated, kairouz2021advances, tran2025privacy}. Each client therefore observes only a partial and potentially biased view of the
target distribution, making centralized continual test-time prompt tuning impractical. Federated Learning (FL) enables clients to collaborate by exchanging
model updates rather than raw data \cite{mcmahan2017communication}. Classical approaches such as FedAvg
\cite{mcmahan2017communication} and FedProx \cite{li2020federated} learn shared models through parameter aggregation, while personalized methods such
as FedAMP \cite{huang2021personalized} and pFedGraph \cite{ye2023personalized} account for client heterogeneity. However, these methods have been studied primarily
with labeled and relatively stationary client data. Recent Federated Test-Time Adaptation (FedTTA) methods \cite{rajib2025fedctta, tan2023heterogeneity}
focus on vision-only models and typically aggregate full model parameters or normalization statistics.
COTE \cite{zhu2026controllable} extends federated prompt learning to test-time adaptation for VLMs, but its adaptation starts from prompts
pretrained on source-domain data. Thus, source-free federated test-time prompt tuning, where adaptation begins without access to source data, remains largely unexplored.

We study Source-Free Federated Test-Time Prompt Tuning (SF-FTTPT), a setting in which distributed clients collaboratively adapt soft prompts on unlabeled test streams
without access to source data. Each client performs local test-time prompt tuning with a frozen CLIP backbone and shares
only lightweight prompt parameters, preserving privacy and communication efficiency. Since the server only aggregates prompt parameters,
any federated method can be combined with any local test-time prompt tuning method. Using this property, we analyze
seven existing federated methods in conjunction with four local test-time prompt tuning methods, namely TPT \cite{shu2022test},
DART \cite{liu2024dart}, SwapPrompt \cite{ma2023swapprompt}, and AdaPrompt \cite{zhang2024robust}, under both IID and non-IID client partitions.
To the best of our knowledge, this is the first such analysis in the source-free test-time setting. Our analysis reveals two key findings. First, under the non-IID partition,
with stable local adaptation methods such as SwapPrompt and AdaPrompt, existing federated methods provide little or no improvement over independent client adaptation. Second,
the benefit of federation strongly depends on the local adaptation method: text prompts can be aggregated effectively across clients, whereas the jointly tuned visual and
class-specific text prompts of DART degrade under prompt sharing.

Extending federated learning to source-free test-time prompt tuning introduces several challenges. First, neither the server nor the clients have access to labels,
so useful client updates must be assessed entirely from unlabeled test-time signals. Second, under non-IID client partitions, strong label shift can cause different clients to
specialize toward different regions of the label space \cite{lu2024federated, wang2020tackling}, making naive aggregation harmful when local adaptations are incompatible.
Third, test-time adaptation is inherently noisy because prompts are updated from small unlabeled mini-batches, so individual updates may be unreliable or biased toward
the current batch. Existing federated methods weight clients either equally or by similarity alone, so they cannot distinguish useful updates from unstable ones,
which causes prompt drift and can destroy client-specific specialization. As our analysis shows, this limits their benefit over independent adaptation, particularly under the non-IID partition.

To address these challenges, we propose FedRAPP (Federated Reliability-Aware Prompt Personalization), a personalized federated prompt aggregation strategy designed
for the source-free test-time setting. FedRAPP has three complementary components. First, it estimates a reliability score for each client's prompt update from
label-free signals, namely the prediction entropy before and after local adaptation and the magnitude of the prompt update. Updates that reduce
uncertainty while remaining stable receive higher reliability. Second, the server keeps an EMA prompt memory for each client with a reliability-dependent coefficient,
so that confident and stable updates are incorporated faster, while noisy mini-batch updates are smoothed. Third, FedRAPP computes personalized collaboration
weights from both the similarity of prompt update directions and the reliability of each source client, so that each client draws mainly from compatible and
reliable collaborators rather than from all clients equally. FedRAPP requires only two additional scalars per client upload and no labels or source data.
As shown in Figure~\ref{fig:intro-radar}, with AdaPrompt as the local adaptation method under the non-IID partition, FedRAPP is the only federated method that
improves over independent adaptation on all nine CLIP datasets.
\textbf{Our contributions are summarized as follows:}

\begin{itemize}
\item We formulate and systematically study \textbf{Source-Free Federated Test-Time Prompt Tuning (SF-FTTPT)}, where distributed clients
collaboratively adapt soft prompts on unlabeled test streams without access to the source data, and establish a
benchmark of seven federated methods and four local adaptation methods under IID and non-IID client partitions, in which every federated method is
applied in a local-adaptation-agnostic manner.

\item Through this analysis, we show that under heterogeneous client distributions, existing federated methods provide little or no improvement
over independent client adaptation, and that the benefit of federation strongly depends on the local adaptation method.

\item We propose \textbf{FedRAPP}, a reliability-aware personalized aggregation framework that combines client reliability estimation, EMA-based prompt memory,
and similarity-guided collaboration. With both SwapPrompt and AdaPrompt, FedRAPP achieves the best average accuracy on nine CLIP datasets under both partitions
and on five ImageNet variants under the non-IID partition, at a small communication and computation overhead.

\end{itemize}

\section{Related Work}

\subsection{Test-Time Prompt Tuning}

Vision-language models such as CLIP \cite{radford2021learning} are effective for zero-shot image recognition, but their performance can degrade when inference prompts are mismatched with the target distribution. Test-Time Adaptation (TTA) methods for vision models \cite{wang2020tent, niu2022efficient, wang2022continual, sun2020test, liang2020we} address distribution shift by adapting model parameters during inference using signals such as entropy and pseudo-labels. However, adapting VLM parameters is costly and may induce catastrophic forgetting \cite{shu2022test}. Prompt tuning is a lightweight alternative that optimizes learnable context vectors while keeping the backbone frozen; CoOp \cite{zhou2022learning}, CoCoOp \cite{zhou2022conditional}, and PLOT \cite{chen2022plot} show its effectiveness but require labeled training data.

Test-time prompt tuning removes this supervision requirement by adapting prompts from unlabeled test data. TPT \cite{shu2022test} minimizes prediction entropy across augmented views, while DiffTPT \cite{feng2023diverse}, C-TPT \cite{yoon2024c}, O-TPT \cite{sharifdeen2025tpt}, and RLCF \cite{zhao2024test} improve augmentation diversity, calibration, or label-free adaptation signals. These methods are typically episodic and do not retain knowledge across the test stream, which continual approaches address: AdaPrompt \cite{zhang2024robust} uses a confidence buffer and prompt ensemble, SwapPrompt \cite{ma2023swapprompt} uses knowledge distillation, DART \cite{liu2024dart} jointly adapts a visual prompt and class-specific text prompts, and DynaPrompt \cite{xiao2025dynaprompt} maintains dynamic prompt buffers. However, all these approaches assume that the test stream is available on a single machine.

\subsection{Federated Learning}

Federated Learning (FL) enables distributed clients to collaboratively learn by exchanging parameters rather than data \cite{mcmahan2017communication}. FedAvg aggregates local models through parameter averaging, but heterogeneous client distributions can cause client drift \cite{zhang2021federated, hsieh2020non}. FedAvgM \cite{hsu2019measuring} adds server-side momentum to stabilize aggregation under non-IID data, FedProx \cite{li2020federated} anchors local updates to the global model, and SCAFFOLD \cite{karimireddy2020scaffold} explicitly corrects drift. Personalized FL methods instead allow client-specific solutions under heterogeneous data \cite{chen2023bayesian, fallah2020personalized, oh2021fedbabu, t2020personalized}; Ditto \cite{li2021ditto} combines global and personalized models, FedAMP \cite{huang2021personalized} promotes selective collaboration among similar clients, and pFedGraph \cite{ye2023personalized} models collaboration through a graph.

In the VLM setting, PromptFL \cite{guo2023promptfl} enables federated CLIP prompt learning, pFedMoAP \cite{luo2025mixture} personalizes it with a mixture of prompt experts, and \cite{tran2025privacy} adds privacy protection for multimodal LLMs, but these approaches remain supervised and rely on labeled or source-domain data. For Federated Test-Time Adaptation (FedTTA), ATP \cite{bao2023adaptive} and FedICON \cite{tan2023heterogeneity} address heterogeneous test-time adaptation, FedCTTA \cite{rajib2025fedctta} introduces similarity-aware aggregation for continual test-time shifts, and BPFedCTTA \cite{wang2026towards} uses Bayesian prior-guided local adaptation and uncertainty-gated aggregation under sequential test-time shifts. However, these methods are designed for standard vision models and aggregate full model parameters or normalization statistics rather than soft prompts. Although Latte \cite{bao2025latte} and COTE \cite{zhu2026controllable} adopt VLM-based federated learning, Latte shares class prototypes across clients, raising privacy concerns, while COTE relies on supervised source pretraining to initialize prompts before test-time adaptation. This leaves source-free federated test-time prompt tuning under heterogeneous client distributions largely unexplored.

\begin{figure*}[t]
    \centering
    \includegraphics[width=0.86\linewidth]{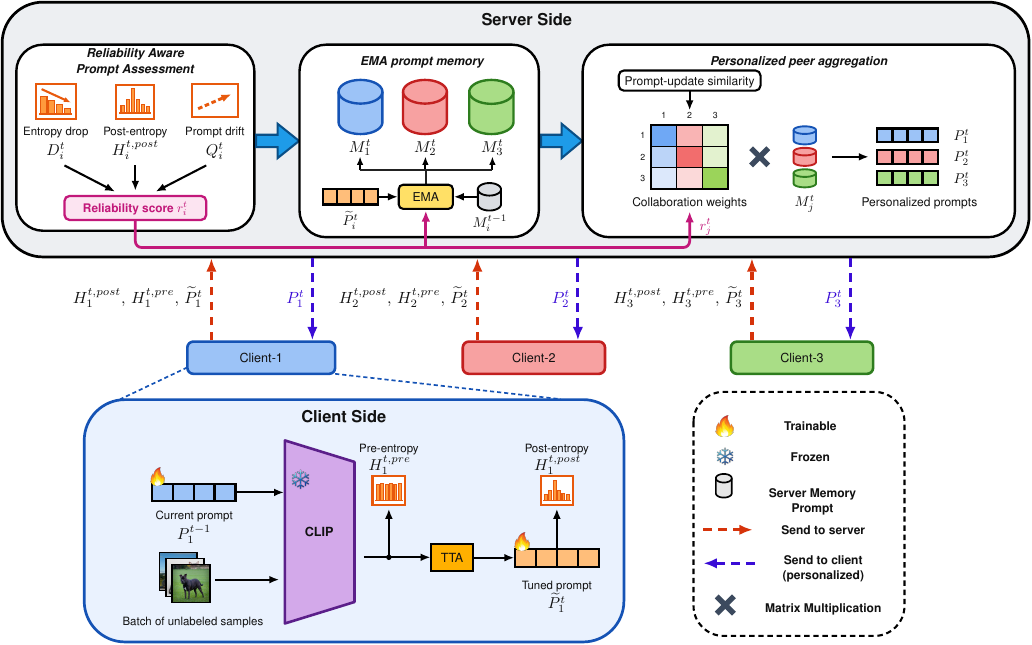}
    \vspace{-0.2em}
    \caption{Working mechanism of FedRAPP (3 clients shown). Clients locally adapt the received prompts and upload the tuned prompt and entropy statistics; the server computes reliability scores, updates the EMA memory, and distributes the personalized prompts to the clients.}
    \label{fig:overview}
\end{figure*}

\section{Methodology}
\label{sec:method}

In federated test-time prompt tuning, each client adapts its prompt using only its unlabeled local samples, so the server cannot rely on labels or validation accuracy to determine which client updates are useful. FedRAPP is a personalized aggregation strategy that allows clients to benefit from compatible collaborators while preserving local test-time specialization. It aggregates prompts by considering two label-free signals: how similarly clients adapt their prompts, and how reliable each client's local adaptation appears from unlabeled test-time statistics.

\subsection{Problem Setup}

We study source-free federated test-time prompt tuning for the vision-language model CLIP, where a set of clients collaboratively adapt soft prompts on unlabeled test streams. Let there be $K$ clients. At communication round $t$, client $i$ receives a mini-batch of unlabeled test samples $\mathcal{B}_i^t = \{x_{i,1}^t, \ldots, x_{i,n_i}^t\}$. The CLIP image encoder and text encoder remain frozen throughout adaptation; only the prompt parameters are updated and communicated.

Each client receives a prompt state $P_i^{t-1}$ from the server at the end of the previous round. In round $t$, client $i$ adapts this prompt on its own unlabeled batch with a local test-time prompt tuning method $\mathrm{TTA}(\cdot)$, which produces a locally adapted prompt $\widetilde{P}_i^t = \mathrm{TTA}(P_i^{t-1}; \mathcal{B}_i^t)$. The server then receives prompt updates from the clients and produces a personalized prompt $P_i^t$ for each client.

\subsection{Local Client Adaptation}

Before tuning the prompt, client $i$ evaluates the current prompt $P_i^{t-1}$ on its unlabeled batch $\mathcal{B}_i^t$ and computes the average prediction entropy, which we term as pre-entropy:
\begin{equation}
H_i^{t,\mathrm{pre}}
=
\frac{1}{n_i}
\sum_{x \in \mathcal{B}_i^t}
H\big(p(y \mid x; P_i^{t-1})\big).
\label{eq:entropy-pre}
\end{equation}
After adapting the prompt using the local test-time prompt tuning method, the client obtains the tuned prompt $\widetilde{P}_i^t$. It then evaluates the tuned prompt on the same batch and computes the post-entropy:
\begin{equation}
H_i^{t,\mathrm{post}}
=
\frac{1}{n_i}
\sum_{x \in \mathcal{B}_i^t}
H\big(p(y \mid x; \widetilde{P}_i^t)\big).
\label{eq:entropy-post}
\end{equation}
The decrease from $H_i^{t,\mathrm{pre}}$ to $H_i^{t,\mathrm{post}}$ provides a label-free signal of whether the local update made the prompt more confident on the current test batch. The client sends the locally tuned prompt and entropy statistics $(\widetilde{P}_i^t, H_i^{t,\mathrm{pre}}, H_i^{t,\mathrm{post}})$ to the server.

\subsection{Reliability-Aware Prompt Assessment}

The server estimates how useful each client update is for collaboration after receiving the tuned prompts and the entropy statistics. It first computes the prompt update direction for each client:
\begin{equation}
\Delta_i^t = \widetilde{P}_i^t - P_i^{t-1}.
\label{eq:prompt-update}
\end{equation}
We later use this update direction to identify clients whose prompts adapt in similar directions. The server further computes the prompt drift:
\begin{equation}
Q_i^t = \log \left(1 + \|\Delta_i^t\|_2\right).
\label{eq:prompt-drift}
\end{equation}
This term measures how far the prompt moves during local adaptation. A large drift may indicate unstable or overly specialized adaptation, so FedRAPP uses it as a penalty when estimating reliability. Next, the server measures how much the prediction entropy decreases after local adaptation. We define the entropy drop as
\begin{equation}
D_i^t = H_i^{t,\mathrm{pre}} - H_i^{t,\mathrm{post}}.
\label{eq:entropy-drop}
\end{equation}
A larger $D_i^t$ indicates that the adapted prompt became more confident on the current batch. To place these quantities on a comparable scale, we normalize the drift, entropy drop, and post-adaptation entropy across the active clients in the current round using
\begin{equation}
z(a_i) = \frac{a_i - \mu(a)}{\sigma(a) + \epsilon},
\label{eq:z-normalization}
\end{equation}
where $\mu(a)$ and $\sigma(a)$ are the mean and standard deviation over the clients, and $\epsilon$ is a small constant for numerical stability.

Using the entropy drop in Eq.~\eqref{eq:entropy-drop}, post-adaptation entropy in Eq.~\eqref{eq:entropy-post}, and prompt drift in Eq.~\eqref{eq:prompt-drift}, the reliability score of each client $i$ is computed as
\begin{equation}
r_i^t
=
z(D_i^t)
-
\lambda_{\mathrm{ent}} z(H_i^{t,\mathrm{post}})
-
\lambda_{\mathrm{drift}} z(Q_i^t).
\label{eq:reliability}
\end{equation}
The first term rewards clients whose adaptation reduces entropy on the current batch, while the second term applies an uncertainty penalty controlled by $\lambda_{\mathrm{ent}}$ for prompts that remain uncertain after adaptation. The final term applies a drift penalty controlled by $\lambda_{\mathrm{drift}}$ for excessive prompt movement. Therefore, higher $r_i^t$ corresponds to a client update that is both confident and stable, making it more suitable for influencing other clients during aggregation.

\subsection{Server-Side EMA Prompt Memory}

Since each client updates its prompt from a small unlabeled batch, a single-round update may be overly influenced by the particular samples in that mini-batch. To reduce the influence of such noisy updates, FedRAPP maintains a server-side exponential moving average (EMA) memory $M_i^t$ for each client, which provides a smoothed representation of the client's recent prompt updates.

We denote $\mu_i^t$ as the EMA coefficient that controls how strongly the previous memory is retained for client $i$ at round $t$. FedRAPP makes this coefficient reliability-dependent, so that stable and confident client updates are incorporated more quickly, while uncertain or excessively drifting updates are smoothed more conservatively:
\begin{equation}
\mu_i^t = \mu_0 - \eta r_i^t.
\label{eq:ema-coefficient}
\end{equation}
Here, $\mu_0$ is the base EMA coefficient and $\eta$ controls the sensitivity of the memory update to the reliability score. In practice, $\mu_i^t$ is bounded to the interval $[0,1]$. The server then updates the prompt memory of client $i$ with the newly received adapted prompt:
\begin{equation}
M_i^t
=
\mu_i^t M_i^{t-1}
+
(1-\mu_i^t)\widetilde{P}_i^t.
\label{eq:ema-memory}
\end{equation}
\subsection{Personalized Peer Aggregation}

For each target client $i$, FedRAPP computes collaboration weights over the active clients that favor source clients with compatible local adaptation. We first compare the prompt update directions of target client $i$ and source client $j$:
\begin{equation}
s_{ij}^{\Delta}
=
\cos(\Delta_i^t, \Delta_j^t).
\label{eq:update-similarity}
\end{equation}
A high similarity indicates that the two clients require compatible prompt adjustments, even though their data remains local and unlabeled, so client $j$ can provide useful information for client $i$.

Since update similarity does not indicate whether a source client's update is trustworthy, FedRAPP also adds the source client's reliability from Eq.~\eqref{eq:reliability} to form the collaboration score:
\begin{equation}
a_{ij}^t = s_{ij}^{\Delta} + r_j^t.
\label{eq:collaboration-score}
\end{equation}
This score favors clients whose prompts adapt in compatible directions and whose local updates are estimated to be reliable. The personalized aggregation weights are obtained by applying a softmax over the collaboration scores:
\begin{equation}
w_{ij}^t
=
\frac{\exp(a_{ij}^t / \tau)}
{\sum_{\ell=1}^{K} \exp(a_{i\ell}^t / \tau)},
\label{eq:collaboration-weights}
\end{equation}
where $\tau$ is a temperature parameter. Using the weights in Eq.~\eqref{eq:collaboration-weights}, FedRAPP then forms a personalized peer prompt for client $i$ from the EMA memory bank:
\begin{equation}
P_{i,\mathrm{peer}}^t
=
\sum_{j=1}^{K} w_{ij}^t M_j^t.
\label{eq:peer-prompt}
\end{equation}
Using the memory bank rather than only the current-round prompts makes the peer prompt less sensitive to noisy updates from a single mini-batch.

Finally, FedRAPP combines the client's own locally adapted prompt with the personalized peer prompt:
\begin{equation}
P_i^t
=
\lambda_{\mathrm{self}} \widetilde{P}_i^t
+
(1-\lambda_{\mathrm{self}}) P_{i,\mathrm{peer}}^t.
\label{eq:personalized-prompt}
\end{equation}
A larger personalization coefficient $\lambda_{\mathrm{self}}$ preserves more of the client's local specialization, while a smaller value allows stronger collaboration.

\section{Experimental Setup}
\begin{table}[t]
\centering
\caption{Average top-1 accuracy over the nine CLIP classification datasets under IID and non-IID partitions. \textbf{Best} and \underline{second-best} per column. Per-dataset results are in Tables~\ref{tab:clip-full-part1} and \ref{tab:clip-full-part2} of the supplementary.}
\label{tab:clip-summary}
\footnotesize
\setlength{\tabcolsep}{2.5pt}
\renewcommand{\arraystretch}{1.0}
\begin{adjustbox}{max width=\columnwidth}
\begin{tabular}{@{}lcccccccc@{}}
\toprule
\multirow{2}{*}{Method}
& \multicolumn{2}{c}{TPT}
& \multicolumn{2}{c}{DART}
& \multicolumn{2}{c}{SwapPrompt}
& \multicolumn{2}{c}{AdaPrompt} \\
\cmidrule(lr){2-3}
\cmidrule(lr){4-5}
\cmidrule(lr){6-7}
\cmidrule(lr){8-9}
& IID & Non-IID
& IID & Non-IID
& IID & Non-IID
& IID & Non-IID \\
\midrule
None-Fed  & 14.87 & 19.15 & \textbf{65.69} & \textbf{66.04} & 65.95 & 66.38 & 66.83 & \underline{66.51} \\
FedAvg    & 27.30 & 30.35 & 26.09 & 27.79 & 66.18 & 65.67 & \underline{66.99} & 65.21 \\
FedAvgM  & 27.88 & 30.99 & 26.20 & 28.38 & 66.21 & 66.50 & 66.64 & 66.48 \\
FedProx   & \textbf{28.37} & 31.64 & 26.27 & 27.16 & \underline{66.31} & 66.08 & 66.49 & 65.35 \\
FedAMP    & 26.34 & 31.95 & \underline{36.03} & 36.38 & 66.05 & \underline{66.55} & 65.25 & 66.46 \\
pFedGraph & 26.91 & \underline{32.99} & 34.67 & 35.20 & 66.23 & 66.26 & 66.39 & 65.48 \\
FedCTTA   & 26.24 & 32.21 & 34.36 & 35.58 & 65.28 & 65.77 & 65.35 & 65.69 \\
pFedMoAP  & 23.86 & 30.26 & 31.61 & 32.64 & 62.43 & 62.82 & 62.89 & 63.73 \\
FedRAPP   & \underline{28.09} & \textbf{34.31} & 36.00 & \underline{36.87} & \textbf{66.77} & \textbf{67.07} & \textbf{67.33} & \textbf{67.68} \\
\bottomrule
\end{tabular}
\end{adjustbox}
\end{table}

\begin{table*}[htbp]
\centering
\caption{Per-dataset non-IID top-1 accuracy with AdaPrompt local adaptation. We report the mean top-1 accuracy $\pm$ standard deviation over three random seeds. \avg{} is the mean over the nine datasets. \textbf{Best} and \underline{second-best} per column.}
\label{tab:clip-path-adaprompt}
\footnotesize
\setlength{\tabcolsep}{2.6pt}
\renewcommand{\arraystretch}{1.0}
\begin{adjustbox}{max width=\textwidth}
\begin{tabular}{@{}lccccccccc@{\hspace{8pt}}c@{}}
\toprule
Method
& Flower102 & Caltech101 & OxfordPets & DTD & Food101 & UCF101 & SUN397 & FGVCAircraft & StanfordCars & \avg \\
\midrule
None-Fed  & \underline{73.11\,$\pm$\,0.13} & 94.02\,$\pm$\,0.04 & 86.11\,$\pm$\,0.08 & 44.40\,$\pm$\,0.11 & \underline{82.35\,$\pm$\,0.07} & 66.29\,$\pm$\,0.09 & 65.14\,$\pm$\,0.10 & 23.16\,$\pm$\,0.13 & 64.01\,$\pm$\,0.08 & \underline{66.51} \\
FedAvg    & 72.01\,$\pm$\,0.19 & 94.00\,$\pm$\,0.22 & \underline{87.22\,$\pm$\,0.27} & \underline{44.56\,$\pm$\,0.08} & 77.54\,$\pm$\,0.13 & 65.02\,$\pm$\,0.10 & 60.19\,$\pm$\,0.41 & 22.88\,$\pm$\,0.09 & 63.48\,$\pm$\,0.33 & 65.21 \\
FedAvgM  & 70.61\,$\pm$\,0.18 & \underline{95.26\,$\pm$\,0.21} & 86.58\,$\pm$\,0.25 & 44.03\,$\pm$\,0.09 & 81.75\,$\pm$\,0.12 & \underline{66.98\,$\pm$\,0.08} & 64.80\,$\pm$\,0.38 & 22.72\,$\pm$\,0.10 & \underline{65.57\,$\pm$\,0.29} & 66.48 \\
FedProx   & 71.49\,$\pm$\,0.10 & 93.83\,$\pm$\,0.33 & 87.09\,$\pm$\,0.15 & 44.30\,$\pm$\,0.25 & 77.51\,$\pm$\,0.06 & 66.93\,$\pm$\,0.30 & 60.33\,$\pm$\,0.25 & 23.01\,$\pm$\,0.12 & 63.64\,$\pm$\,0.16 & 65.35 \\
FedAMP    & 72.30\,$\pm$\,0.38 & 94.95\,$\pm$\,0.18 & 85.29\,$\pm$\,0.13 & 44.41\,$\pm$\,0.13 & 82.11\,$\pm$\,0.37 & 66.96\,$\pm$\,0.12 & \underline{65.18\,$\pm$\,0.14} & 22.99\,$\pm$\,0.31 & 63.96\,$\pm$\,0.42 & 66.46 \\
pFedGraph & 72.07\,$\pm$\,0.36 & 93.65\,$\pm$\,0.04 & 86.75\,$\pm$\,0.18 & 44.51\,$\pm$\,0.22 & 80.23\,$\pm$\,0.36 & 64.59\,$\pm$\,0.12 & 60.69\,$\pm$\,0.32 & \underline{23.21\,$\pm$\,0.06} & 63.61\,$\pm$\,0.37 & 65.48 \\
FedCTTA   & 72.12\,$\pm$\,0.16 & 94.67\,$\pm$\,0.18 & 85.01\,$\pm$\,0.12 & 44.11\,$\pm$\,0.12 & 82.00\,$\pm$\,0.14 & 64.82\,$\pm$\,0.12 & 62.56\,$\pm$\,0.21 & 20.94\,$\pm$\,0.27 & 64.93\,$\pm$\,0.20 & 65.69 \\
pFedMoAP  & 68.37\,$\pm$\,0.37 & 91.74\,$\pm$\,0.19 & 85.39\,$\pm$\,0.21 & 41.57\,$\pm$\,0.28 & 79.75\,$\pm$\,0.41 & 62.82\,$\pm$\,0.38 & 61.72\,$\pm$\,0.30 & 19.67\,$\pm$\,0.27 & 62.55\,$\pm$\,0.28 & 63.73 \\
FedRAPP   & \textbf{73.79\,$\pm$\,0.07} & \textbf{95.92\,$\pm$\,0.13} & \textbf{89.18\,$\pm$\,0.05} & \textbf{45.11\,$\pm$\,0.11} & \textbf{82.44\,$\pm$\,0.13} & \textbf{67.73\,$\pm$\,0.12} & \textbf{65.52\,$\pm$\,0.10} & \textbf{23.25\,$\pm$\,0.11} & \textbf{66.17\,$\pm$\,0.09} & \textbf{67.68} \\
\bottomrule
\end{tabular}
\end{adjustbox}
\end{table*}

\begin{table*}[t]
\centering
\caption{Top-1 accuracy (mean\,$\pm$\,standard deviation) on ImageNet and its variants under IID and non-IID partitions for AdaPrompt and SwapPrompt. \textbf{Best} and \underline{second-best} per column. The corresponding TPT and DART results are in Table~\ref{tab:imagenet-tpt-dart} of the supplementary.}
\label{tab:imagenet-variants}
\scriptsize
\setlength{\tabcolsep}{1.8pt}
\renewcommand{\arraystretch}{1.0}
\begin{adjustbox}{max width=\textwidth}
\begin{tabular}{@{}ll*{12}{c}@{}}
\toprule
\multirow{2}{*}{TTA}
& \multirow{2}{*}{Fed.}
& \multicolumn{2}{c}{ImageNet}
& \multicolumn{2}{c}{ImageNet-V2}
& \multicolumn{2}{c}{ImageNet-A}
& \multicolumn{2}{c}{ImageNet-S}
& \multicolumn{2}{c}{ImageNet-R}
& \multicolumn{2}{c}{Average} \\
\cmidrule(lr){3-4}
\cmidrule(lr){5-6}
\cmidrule(lr){7-8}
\cmidrule(lr){9-10}
\cmidrule(lr){11-12}
\cmidrule(lr){13-14}
& & IID & Non-IID & IID & Non-IID & IID & Non-IID & IID & Non-IID & IID & Non-IID & IID & Non-IID \\
\midrule
\multirow{9}{*}{SwapPrompt}
& None-Fed  & 65.41\,$\pm$\,0.06 & \underline{65.60\,$\pm$\,0.08} & \textbf{60.17\,$\pm$\,0.13} & 60.03\,$\pm$\,0.14 & \underline{49.29\,$\pm$\,0.09} & 49.65\,$\pm$\,0.08 & 48.01\,$\pm$\,0.10 & 48.00\,$\pm$\,0.09 & 73.73\,$\pm$\,0.09 & 73.91\,$\pm$\,0.09 & 59.32 & 59.44 \\
& FedAvg    & 65.41\,$\pm$\,0.26 & 65.31\,$\pm$\,0.36 & 60.15\,$\pm$\,0.21 & 59.92\,$\pm$\,0.07 & \textbf{49.31\,$\pm$\,0.13} & 49.52\,$\pm$\,0.04 & \textbf{49.02\,$\pm$\,0.36} & 47.60\,$\pm$\,0.07 & \textbf{74.65\,$\pm$\,0.06} & 73.33\,$\pm$\,0.22 & \textbf{59.71} & 59.14 \\
& FedAvgM  & 65.43\,$\pm$\,0.24 & 65.40\,$\pm$\,0.31 & \underline{60.16\,$\pm$\,0.18} & 59.93\,$\pm$\,0.06 & 48.01\,$\pm$\,0.12 & 48.80\,$\pm$\,0.03 & 48.93\,$\pm$\,0.31 & 48.27\,$\pm$\,0.05 & 73.81\,$\pm$\,0.03 & 73.95\,$\pm$\,0.19 & 59.27 & 59.27 \\
& FedProx   & 65.41\,$\pm$\,0.35 & 65.46\,$\pm$\,0.18 & 59.73\,$\pm$\,0.34 & 59.90\,$\pm$\,0.10 & 49.22\,$\pm$\,0.20 & 49.54\,$\pm$\,0.20 & 48.99\,$\pm$\,0.12 & 48.77\,$\pm$\,0.30 & 74.46\,$\pm$\,0.16 & 74.57\,$\pm$\,0.07 & 59.56 & 59.65 \\
& FedAMP    & \textbf{65.60\,$\pm$\,0.20} & 65.42\,$\pm$\,0.28 & 59.90\,$\pm$\,0.21 & \underline{60.05\,$\pm$\,0.20} & 49.21\,$\pm$\,0.24 & \textbf{50.63\,$\pm$\,0.05} & \underline{49.00\,$\pm$\,0.17} & \underline{49.02\,$\pm$\,0.33} & \underline{74.58\,$\pm$\,0.33} & \underline{75.00\,$\pm$\,0.00} & \underline{59.66} & \underline{60.02} \\
& pFedGraph & 65.41\,$\pm$\,0.11 & 65.48\,$\pm$\,0.18 & 59.79\,$\pm$\,0.29 & 59.73\,$\pm$\,0.33 & 49.22\,$\pm$\,0.15 & 49.53\,$\pm$\,0.19 & 48.87\,$\pm$\,0.12 & 48.75\,$\pm$\,0.43 & 74.27\,$\pm$\,0.26 & 74.49\,$\pm$\,0.16 & 59.51 & 59.60 \\
& FedCTTA   & 63.69\,$\pm$\,0.08 & 64.29\,$\pm$\,0.09 & 58.58\,$\pm$\,0.10 & 58.85\,$\pm$\,0.15 & 48.03\,$\pm$\,0.16 & 48.71\,$\pm$\,0.10 & 47.91\,$\pm$\,0.10 & 48.18\,$\pm$\,0.11 & 73.27\,$\pm$\,0.07 & 73.99\,$\pm$\,0.15 & 58.30 & 58.80 \\
& pFedMoAP  & 62.22\,$\pm$\,0.41 & 61.88\,$\pm$\,0.37 & 56.48\,$\pm$\,0.37 & 56.77\,$\pm$\,0.36 & 45.43\,$\pm$\,0.47 & 46.07\,$\pm$\,0.31 & 45.64\,$\pm$\,0.27 & 46.81\,$\pm$\,0.40 & 70.99\,$\pm$\,0.27 & 71.90\,$\pm$\,0.31 & 56.15 & 56.69 \\
& FedRAPP   & \underline{65.59\,$\pm$\,0.02} & \textbf{65.68\,$\pm$\,0.04} & 59.99\,$\pm$\,0.11 & \textbf{60.19\,$\pm$\,0.12} & 49.24\,$\pm$\,0.14 & \underline{49.78\,$\pm$\,0.07} & 48.76\,$\pm$\,0.05 & \textbf{49.83\,$\pm$\,0.08} & 74.17\,$\pm$\,0.05 & \textbf{75.39\,$\pm$\,0.10} & 59.55 & \textbf{60.17} \\
\midrule
\multirow{9}{*}{AdaPrompt}
& None-Fed  & 67.87\,$\pm$\,0.07 & \underline{67.71\,$\pm$\,0.10} & 60.87\,$\pm$\,0.14 & \underline{61.07\,$\pm$\,0.11} & 48.68\,$\pm$\,0.07 & 48.83\,$\pm$\,0.05 & 49.71\,$\pm$\,0.06 & 49.40\,$\pm$\,0.08 & 75.46\,$\pm$\,0.09 & 73.68\,$\pm$\,0.08 & 60.52 & 60.14 \\
& FedAvg    & 67.98\,$\pm$\,0.42 & 67.13\,$\pm$\,0.28 & 61.22\,$\pm$\,0.12 & 60.66\,$\pm$\,0.27 & \underline{48.76\,$\pm$\,0.21} & 48.52\,$\pm$\,0.39 & 49.77\,$\pm$\,0.11 & 48.69\,$\pm$\,0.42 & 76.41\,$\pm$\,0.31 & 72.46\,$\pm$\,0.29 & \underline{60.83} & 59.49 \\
& FedAvgM  & \textbf{68.11\,$\pm$\,0.39} & 67.29\,$\pm$\,0.27 & 61.21\,$\pm$\,0.10 & 60.57\,$\pm$\,0.25 & 48.01\,$\pm$\,0.19 & 48.36\,$\pm$\,0.37 & \underline{49.80\,$\pm$\,0.08} & \textbf{49.71\,$\pm$\,0.41} & 76.39\,$\pm$\,0.30 & 71.65\,$\pm$\,0.26 & 60.70 & 59.51 \\
& FedProx   & 68.00\,$\pm$\,0.31 & 67.16\,$\pm$\,0.30 & 60.96\,$\pm$\,0.18 & 60.77\,$\pm$\,0.13 & 48.73\,$\pm$\,0.29 & 48.78\,$\pm$\,0.18 & 49.64\,$\pm$\,0.03 & 48.90\,$\pm$\,0.09 & \underline{76.48\,$\pm$\,0.20} & 73.41\,$\pm$\,0.28 & 60.76 & 59.80 \\
& FedAMP    & 67.08\,$\pm$\,0.28 & 66.97\,$\pm$\,0.18 & \underline{61.51\,$\pm$\,0.22} & 60.72\,$\pm$\,0.39 & 48.59\,$\pm$\,0.35 & 48.87\,$\pm$\,0.21 & 47.65\,$\pm$\,0.10 & 49.43\,$\pm$\,0.10 & 75.93\,$\pm$\,0.33 & \underline{75.82\,$\pm$\,0.21} & 60.15 & \underline{60.36} \\
& pFedGraph & 67.99\,$\pm$\,0.40 & 67.07\,$\pm$\,0.40 & 60.89\,$\pm$\,0.04 & 60.79\,$\pm$\,0.07 & 48.60\,$\pm$\,0.23 & \underline{48.91\,$\pm$\,0.05} & 49.70\,$\pm$\,0.38 & 49.47\,$\pm$\,0.26 & \textbf{76.50\,$\pm$\,0.18} & 72.26\,$\pm$\,0.30 & 60.74 & 59.70 \\
& FedCTTA   & 66.77\,$\pm$\,0.14 & 67.60\,$\pm$\,0.19 & 60.59\,$\pm$\,0.11 & 61.05\,$\pm$\,0.15 & 47.17\,$\pm$\,0.14 & 47.31\,$\pm$\,0.16 & 48.35\,$\pm$\,0.13 & 48.48\,$\pm$\,0.09 & 73.87\,$\pm$\,0.08 & 70.39\,$\pm$\,0.09 & 59.35 & 58.97 \\
& pFedMoAP  & 64.88\,$\pm$\,0.51 & 62.65\,$\pm$\,0.43 & 57.43\,$\pm$\,0.31 & 58.84\,$\pm$\,0.43 & 42.75\,$\pm$\,0.41 & 43.16\,$\pm$\,0.39 & 46.04\,$\pm$\,0.51 & 46.16\,$\pm$\,0.58 & 71.82\,$\pm$\,0.32 & 71.79\,$\pm$\,0.34 & 56.59 & 56.52 \\
& FedRAPP   & \underline{68.02\,$\pm$\,0.11} & \textbf{67.88\,$\pm$\,0.12} & \textbf{62.49\,$\pm$\,0.06} & \textbf{61.92\,$\pm$\,0.11} & \textbf{48.89\,$\pm$\,0.05} & \textbf{48.97\,$\pm$\,0.01} & \textbf{49.88\,$\pm$\,0.09} & \underline{49.68\,$\pm$\,0.02} & 76.19\,$\pm$\,0.02 & \textbf{76.59\,$\pm$\,0.02} & \textbf{61.09} & \textbf{61.01} \\
\bottomrule
\end{tabular}
\end{adjustbox}
\end{table*}

\label{sec:setup}

\paragraph{Datasets}
We conduct experiments on nine standard CLIP image classification benchmark datasets: Flower102, Caltech101, Food101, DTD, OxfordPets, SUN397, StanfordCars, FGVCAircraft, and UCF101. We also evaluate on ImageNet and its variants ImageNetV2, ImageNet-A, ImageNet-Sketch, and ImageNet-R under natural distribution shift. Following the standard continual TTA protocol, the test split of each dataset is used for both test-time adaptation and immediate evaluation.

\paragraph{Implementation Details}
We use CLIP with a frozen ViT-B/16 image encoder and four local test-time prompt tuning methods: TPT \cite{shu2022test}, AdaPrompt \cite{zhang2024robust}, SwapPrompt \cite{ma2023swapprompt}, and DART \cite{liu2024dart}. We compare FedAvg \cite{mcmahan2017communication}, FedAvgM \cite{hsu2019measuring}, FedProx \cite{li2020federated}, FedAMP \cite{huang2021personalized}, pFedGraph \cite{ye2023personalized}, FedCTTA \cite{rajib2025fedctta}, and pFedMoAP \cite{luo2025mixture}, with ``None-Fed'' denoting independent client adaptation. For pFedMoAP, we replace its supervised gating network with parameter-free dot-product attention over the client prompts, since test-time adaptation has no access to labeled data. All main-table experiments (Tables~\ref{tab:clip-summary}--\ref{tab:imagenet-variants}) use $K=10$ clients. Under the IID partition, the images of each
class are split evenly across the clients. Under the pathological non-IID
partition~\cite{guo2023promptfl, luo2025mixture}, the classes are randomly
shuffled and divided evenly into 10 disjoint groups, and each client
receives all test images of one group, so no two clients share a class.
For instance, each client holds 3 to 4 classes on OxfordPets (37 classes) and
20 classes on ImageNet-A (200 classes). In both partitions, every test
image belongs to exactly one client and is used exactly once. Only prompts are tuned and aggregated, with batch size 16. We set $\lambda_{\mathrm{ent}}=0.5$, $\lambda_{\mathrm{drift}}=0.25$, $\lambda_{\mathrm{self}}=0.4$, using OxfordPets, Flower102, and Caltech101 for tuning, and set $\tau=1$, $\epsilon=10^{-12}$, $\mu_0=0.8$, and $\eta=0.1$. Main-table results are the mean and standard deviation over three random seeds.

\section{Result Analysis}
\label{sec:results}

\subsection{Performance Analysis of Federated Aggregation Methods}
\label{sec:fed-analysis}

Table~\ref{tab:clip-summary} reports the average top-1 accuracy over the nine CLIP datasets, Table~\ref{tab:clip-path-adaprompt} the per-dataset non-IID results with AdaPrompt, and Table~\ref{tab:imagenet-variants} the results on ImageNet and its variants with SwapPrompt and AdaPrompt;
the complete results are in Tables~\ref{tab:clip-full-part1}, \ref{tab:clip-full-part2}, and~\ref{tab:imagenet-tpt-dart} of the supplementary material. We use None-Fed as the reference, since a federated method is only beneficial if it improves over independent adaptation.

\paragraph{Federation under the non-IID partition}
Existing federated methods struggle to improve over None-Fed.
With AdaPrompt, every existing Federated method remains below None-Fed on the nine CLIP datasets average (66.51), ranging from
63.73 for pFedMoAP to 66.48 for FedAvgM, whereas FedRAPP reaches 67.68, an improvement of 1.17 over the None-Fed baseline (Table~\ref{tab:clip-summary}).
With SwapPrompt, the strongest baselines, FedAMP and FedAvgM, improve over None-Fed by only 0.17 and 0.12, respectively, while FedRAPP improves by 0.69.
We observe the same trend on the ImageNet variants, as shown in Table~\ref{tab:imagenet-variants}; FedRAPP achieves the best average accuracy for both methods,
improving over None-Fed by 0.73 with SwapPrompt (60.17 vs.\ 59.44) and by 0.87 with AdaPrompt (61.01 vs.\ 60.14), while six of the seven baselines remain below
None-Fed with AdaPrompt. The per-dataset results in Table~\ref{tab:clip-path-adaprompt} show that FedRAPP achieves the best accuracy on all nine datasets with AdaPrompt,
with the largest gains over None-Fed on OxfordPets (+3.07), StanfordCars (+2.16), and Caltech101 (+1.90). In contrast, naive aggregation degrades performance on heterogeneous
datasets; for instance, FedAvg is 4.95 lower than None-Fed on SUN397 and 4.81 lower on Food101. This indicates that reliability-aware collaboration helps avoid
the harmful averaging of incompatible client updates. pFedMoAP, which combines peer prompts as non-local experts, performs worse than None-Fed in this source-free
setting.

\paragraph{Federation under the IID partition}
When the client data are similar, global aggregation is already effective, and the differences between the methods are smaller. FedRAPP still achieves the best
nine-dataset average for both SwapPrompt (66.77) and AdaPrompt (67.33), ahead of the strongest global methods, FedProx (66.31) and FedAvg (66.99), respectively
as shown in Table~\ref{tab:clip-summary}. On the ImageNet variants, FedRAPP also achieves the best IID average with AdaPrompt (61.09, Table~\ref{tab:imagenet-variants}).
This shows that the reliability-aware personalization of FedRAPP retains the benefit of collaboration when the client distributions are similar.

\subsection{Effect of the Local Adaptation Method}
\label{sec:tta-effect}

The choice of local adaptation method strongly influences the effectiveness of federated aggregation:
Table~\ref{tab:clip-summary} shows that TPT and DART are unstable under federation, while SwapPrompt and AdaPrompt are more robust.
First, TPT collapses in the continual test-time setting, as repeated entropy minimization
on the same prompt accumulates errors over the test stream~\cite{xiao2025dynaprompt, zhang2024robust},
reaching only 14.87 and 19.15 under the IID and non-IID partitions without federation.
However, federation recovers a large part of this performance; every federated method improves over None-Fed,
and FedRAPP provides the strongest non-IID improvement, reaching 34.31 (+15.16 over None-Fed).
This shows that aggregating text prompts across clients can push them towards stability even with an unstable local adaptation scheme.
In contrast, DART achieves strong standalone performance (65.69 and 66.04 under the IID and non-IID partitions)
but degrades under prompt collaboration; for instance, FedAvg reduces its accuracy to 26.09 and 27.79, respectively.
DART jointly tunes an image prompt and class-specific text prompts, and these per-class, dual-modal prompts are highly sensitive to the choice of
shared prompt set (Section~\ref{sec:dart-federation} of the supplementary material), so aggregating them disrupts the joint adaptation even when clients share the same classes.
Personalized aggregation partially recovers this performance, with FedRAPP reaching 36.00 and 36.87 for IID and non-IID, respectively,
although it remains well below the standalone setting.
SwapPrompt tunes only text prompts and stabilizes its local adaptation with an EMA-based target prompt and a consistency objective~\cite{ma2023swapprompt},
so it remains robust to federation; with SwapPrompt, FedRAPP improves over None-Fed by 0.82 and 0.69 under the IID and non-IID partitions, respectively.
Finally, AdaPrompt combines an ensemble of three tunable text prompts with a confidence-aware memory~\cite{zhang2024robust} and benefits the most from FedRAPP:
under the non-IID partition, the gain of FedRAPP over None-Fed is larger with AdaPrompt than with SwapPrompt on both the CLIP datasets (1.17 vs.\ 0.69)
and the ImageNet variants (0.87 vs.\ 0.73), and FedRAPP with AdaPrompt achieves the highest overall accuracy (67.68 and 61.01).
This suggests that the ensemble of three text prompts provides richer prompt updates for reliability-aware aggregation.
The TPT and DART results on the ImageNet variants in Table~\ref{tab:imagenet-tpt-dart} of the supplementary material follow the same trends.

\subsection{Ablation Analysis}
All experiments in this section use the non-IID partition with AdaPrompt as the local adaptation method. The component ablation and the hyperparameter sensitivity use OxfordPets, Caltech101, and Flower102.

\begin{figure*}[t]
    \centering
    \includegraphics[width=\textwidth]{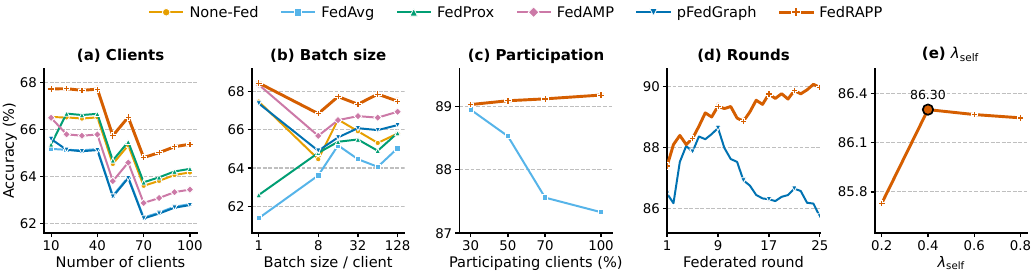}
    \vspace{-1.4em}
    \caption{Sensitivity and scalability under the non-IID partition with AdaPrompt as the local adaptation method. (a) Average Accuracy vs varying number of clients over the nine CLIP datasets. (b) Average Accuracy vs varying batch size over the nine CLIP datasets. (c) Partial participation on OxfordPets. (d) Average Accuracy of all clients vs federated rounds on OxfordPets (e) Average Accuracy vs ~$\lambda_{\mathrm{self}}$ (Personalization Coefficient) across OxfordPets, Flower102, Caltech101.}
    \label{fig:sensitivity}
\end{figure*}

\begin{table}[t]
\centering
\caption{Component ablation of FedRAPP under the non-IID partition with AdaPrompt local adaptation. \textbf{Best} in bold.}
\label{tab:fedrapp-ablation}
\footnotesize
\setlength{\tabcolsep}{4pt}
\renewcommand{\arraystretch}{1.0}
\begin{adjustbox}{max width=\columnwidth}
\begin{tabular}{ccc|cccc}
\toprule
Sim. & Rel. & EMA & OxfordPets & Caltech101 & Flower102 & Average \\
\midrule
\xmark & \xmark & \xmark & 86.89 & 93.96 & 72.12 & 84.32 \\
\cmark & \xmark & \xmark & 87.62 & 94.08 & 72.23 & 84.64 \\
\xmark & \cmark & \xmark & 87.74 & 94.06 & 72.55 & 84.78 \\
\xmark & \xmark & \cmark & 88.01 & 94.29 & 72.62 & 84.97 \\
\cmark & \xmark & \cmark & 88.51 & 94.93 & 72.70 & 85.38 \\
\cmark & \cmark & \xmark & 88.11 & 94.59 & 72.72 & 85.14 \\
\xmark & \cmark & \cmark & 88.67 & 94.96 & 73.09 & 85.57 \\
\cmark & \cmark & \cmark & \textbf{89.18} & \textbf{95.92} & \textbf{73.79} & \textbf{86.30} \\
\bottomrule
\end{tabular}
\end{adjustbox}
\end{table}

\paragraph{Component ablation} Table~\ref{tab:fedrapp-ablation} examines the three FedRAPP components: update similarity (Sim.), reliability (Rel.), and server-side EMA memory (EMA). Removing all three gives the lowest average accuracy (84.32). Among the individual components, EMA gives the largest gain (84.97), followed by reliability (84.78) and update similarity (84.64), and among pairs, Rel.+EMA performs best (85.57). Update similarity gives the smallest gain alone, but adding it to Rel.+EMA raises the average from 85.57 to 86.30, the best result on all three datasets. This shows that the three components are complementary.

\paragraph{Sensitivity to $\lambda_{\mathrm{ent}}$ and $\lambda_{\mathrm{drift}}$} First, we set $\lambda_{\mathrm{self}}$ to 0 to find the best values of $\lambda_{\mathrm{ent}}$ and $\lambda_{\mathrm{drift}}$ used in the reliability score. We sweep both coefficients over $\{0, 0.25, 0.5, 1.0\}$ and monitor the average accuracy. As shown in Table~\ref{tab:lambda-grid}, the best values are $\lambda_{\mathrm{ent}}=0.5$ and $\lambda_{\mathrm{drift}}=0.25$.

\paragraph{Sensitivity to $\lambda_{\mathrm{self}}$} The personalization coefficient controls the balance between the client's own adapted prompt and the personalized peer prompt. With $\lambda_{\mathrm{ent}}$ and $\lambda_{\mathrm{drift}}$ fixed, Figure~\ref{fig:sensitivity}(e) shows that accuracy stays high for $\lambda_{\mathrm{self}}$ between 0.4 and 0.8, with the best value at 0.4 (86.30). Accuracy drops at $\lambda_{\mathrm{self}}=0.2$ (85.73), because a low value relies too much on the peer prompt and reduces the contribution of the client's own adaptation.

\begin{table}[t]
\centering
\caption{Average top-1 accuracy (\%) over OxfordPets, Flower102 and Caltech101 (non-IID) for $\lambda_{\mathrm{ent}}$ and $\lambda_{\mathrm{drift}}$.}
\label{tab:lambda-grid}
\footnotesize
\setlength{\tabcolsep}{5pt}
\renewcommand{\arraystretch}{1.0}
\begin{adjustbox}{max width=\columnwidth}
\begin{tabular}{@{}lcccc@{}}
\toprule
\multirow{2}{*}{$\lambda_{\mathrm{ent}}$} & \multicolumn{4}{c}{$\lambda_{\mathrm{drift}}$} \\
\cmidrule(l{2pt}r{2pt}){2-5}
& 0 & 0.25 & 0.5 & 1.0 \\
\midrule
0     & 84.54 & 84.85 & 84.78 & 84.43 \\
0.25  & 84.97 & 85.20 & 85.09 & 84.72 \\
0.5   & 85.13 & \textbf{85.41} & 85.25 & 84.84 \\
1.0   & 84.75 & 85.02 & 84.93 & 84.57 \\
\bottomrule
\end{tabular}
\end{adjustbox}
\end{table}

\paragraph{Number of clients} In Figure~\ref{fig:sensitivity}(a), we vary the number of clients from 10 to 100. The average accuracy of the federated methods over the nine CLIP datasets remains fairly stable up to around 40 clients and declines as we further increase the number of clients. However, FedRAPP outperforms all the other federated methods with more clients, demonstrating robustness under severe heterogeneity.

\paragraph{Client batch size} In Figure~\ref{fig:sensitivity}(b), we vary the batch size from 1 to 128 on the nine CLIP datasets. FedAvg and FedProx improve as the batch size grows, while the personalized methods, including FedRAPP, perform well even at batch size 1. FedRAPP stays stable across batch sizes because its EMA memory builds a reliable prompt estimate across rounds and reduces the noise of small batches.

\paragraph{Partial client participation} In Figure~\ref{fig:sensitivity}(c), we randomly sample a fraction of the clients in each round on OxfordPets. FedRAPP remains stable at low participation and improves as more clients participate, whereas FedAvg degrades because aggregation becomes noisier.

\paragraph{Stability over federated rounds} Figure~\ref{fig:sensitivity}(d) shows the average client accuracy per round on OxfordPets. The accuracy of FedRAPP increases steadily over the rounds, whereas that of pFedGraph decreases.

\paragraph{Communication and computation overhead} Table~\ref{tab:cost} reports the transmitted parameters and the running time per round. The transmitted parameters are the learnable prompt(s) of the local TTA method: $d=7{,}168$ for AdaPrompt and $d=2{,}048$ for SwapPrompt. In both FedAvg and FedRAPP, each client uploads its prompt(s) and downloads one prompt of the same size: the global average in FedAvg and its personalized prompt in FedRAPP. FedRAPP only adds two entropy scalars to the upload ($0.03\%$ for AdaPrompt, $0.10\%$ for SwapPrompt) and two entropy evaluations on the client ($0.09$\,s of the extra $0.13$\,s), so it takes $1.29$\,s per client per round, $1.12\times$ the time of FedAvg. The server holds no model or data, only one EMA prompt per client ($280$\,KB for $K=10$ with AdaPrompt), and runs only closed-form operations: reliability scores, EMA updates, and pairwise similarities in $\mathcal{O}(K^2 d)$, which take about $10$\,ms per round with 10 clients, less than $1\%$ of the client time.

\begin{table}[t]
\centering
\small
\setlength{\tabcolsep}{4pt}
\caption{Transmitted parameters per client and running time per round. $d$: learnable prompt size of the local TTA method. Time is per client per round, with the factor relative to FedAvg.}
\label{tab:cost}
\begin{tabular}{lccc}
\toprule
Method & Upload & Download & Time / round (s) \\
\midrule
FedAvg   & $d$     & $d$       & 1.15 (1.00$\times$) \\
FedAMP   & $d$     & $d$       & 1.26 (1.10$\times$) \\
FedRAPP  & $d{+}2$ & $d$       & 1.29 (1.12$\times$) \\
\bottomrule
\end{tabular}
\end{table}

\section{Conclusion}
In this work, we propose FedRAPP, a reliability-aware personalized aggregation framework for source-free federated test-time prompt tuning of vision-language models.
Through reliability-guided client assessment, EMA-based prompt memory, and similarity-aware collaboration, FedRAPP effectively preserves local prompt specialization while
enabling beneficial knowledge sharing among distributed clients at test time. Our analysis of seven existing federated methods with four local test-time prompt
tuning methods showed that existing methods provide limited benefit over independent client adaptation, particularly under non-IID partitions, and this benefit
strongly depends on the local adaptation method. Experimental results across nine CLIP benchmark datasets and five ImageNet variants
showed that FedRAPP outperforms existing federated methods, particularly under the non-IID partition, at a small communication and computation overhead. The ablation and sensitivity analyses further showed that the three components are complementary and that FedRAPP is robust to the number of clients,
batch size, client participation, and federated rounds. These findings suggest that personalized
and reliability-aware collaboration is essential for effective SF-FTTPT of vision-language models.

{
    \small
    \bibliographystyle{ieeenat_fullname}
    \bibliography{egbib}
}

\clearpage
\maketitlesupplementary
\setcounter{section}{0}
\setcounter{table}{0}
\setcounter{figure}{0}
\renewcommand{\thesection}{\Alph{section}}
\renewcommand{\thetable}{A\arabic{table}}
\renewcommand{\thefigure}{A\arabic{figure}}
\renewcommand{\theHsection}{supp.\arabic{section}}
\renewcommand{\theHtable}{supp.\arabic{table}}
\renewcommand{\theHfigure}{supp.\arabic{figure}}

\section{Additional Implementation Details}
\label{sec:impl-details}
\noindent\textbf{Prompt templates.}
TPT, SwapPrompt and DART initialize the learnable prompt from the template
``a photo of a [CLASS].'', whereas AdaPrompt uses an ensemble of three learnable
prompts, initialized from the templates
``an image of a [CLASS].'',
``a colorful image of a [CLASS].'' and
``a noisy picture of a [CLASS].''
For all methods, only the context tokens of the text prompts are tuned at test
time, following each TTA method's implementation. DART additionally tunes an
image prompt of length 2, which is added to the second layer of the CLIP image encoder
and initialized from a uniform distribution on $(-1, 1)$, as in the original
work~\cite{liu2024dart}.

\noindent\textbf{Federated protocol.}
In this Federated Test-time Adaptation setup, all clients start from the same initial prompt(s) as described
above and adapt only on their own unlabeled test images. The test images of each client are split into
batches of 16, and one batch is processed in each communication round. In each round,
(i) each client adapts its current prompt(s) on its batch with the local TTA method,
(ii) uploads the adapted prompt(s) to the server,
(iii) the server aggregates them and sends one prompt back to each client, and
(iv) each client predicts the labels of the same batch with the received prompt, which is also its starting
point for the next round.
The server aggregates in every round; therefore, each client
communicates once per 16 test images.

\section{Performance analysis on CLIP Benchmark datasets}

Table~\ref{tab:clip-full-part1} and Table~\ref{tab:clip-full-part2} show the detailed performance comparison of different federated learning methods combined with multiple test-time adaptation (TTA) techniques under both IID and non-IID data partitions across the CLIP benchmark datasets. Overall, the proposed FedRAPP framework demonstrates strong and consistent performance across most datasets and partition settings, particularly under challenging non-IID scenarios.

For the TPT setting, FedRAPP achieves the best performance in most non-IID cases, including Flower102, Caltech101, OxfordPets, DTD, and SUN397, indicating that the proposed method effectively improves robustness under heterogeneous client distributions. Although FedProx and FedAvgM obtain higher scores on a few datasets, such as Food101, UCF101, FGVCAircraft, and StanfordCars, FedRAPP consistently provides strong adaptation under distribution shifts. pFedMoAP, which combines peer prompts as non-local experts, shows lower performance across datasets in this source-free setting; an analysis of its original attention-gated variant is provided in Section~\ref{sec:pfedmoap-gating}. TPT in the continual setup performs poorly across all the datasets. However, federated aggregation improves the performance of TPT-continual, showing that collaboration for prompt tuning improves the results. Under the DART setting, the non-federated baseline achieves the highest overall accuracy, suggesting that DART is sensitive to federated optimization. This can be attributed to the fact that DART jointly tunes a visual prompt and class-specific text prompts, which makes it highly sensitive to the choice of shared prompt set (Section~\ref{sec:dart-federation}). Nevertheless, among federated approaches, FedRAPP consistently provides the best or near-best performance, achieving the best federated performance across all the non-IID datasets and demonstrating improved stability compared to existing federated baselines. For SwapPrompt, FedRAPP achieves the best or near-best performance across most non-IID datasets, including Flower102, Caltech101, OxfordPets, DTD, Food101, UCF101, SUN397, and FGVCAircraft. This indicates that FedRAPP effectively preserves prompt adaptation capability even under decentralized heterogeneous data distributions. The improvements are particularly noticeable in Flower102, OxfordPets, Food101, and UCF101 under non-IID partitions. Similarly, under the AdaPrompt setting, FedRAPP consistently achieves the best performance across the datasets, especially in non-IID environments. Under the non-IID partition, FedRAPP obtains the highest scores on all nine datasets, Flower102, Caltech101, OxfordPets, DTD, Food101, UCF101, SUN397, FGVCAircraft, and StanfordCars, highlighting its effectiveness in handling client-specific feature variations and severe data heterogeneity.

Overall, the results demonstrate that the proposed FedRAPP framework provides strong generalization and robustness across diverse datasets and TTA methods. In particular, its consistent improvements under non-IID settings indicate that the method is highly effective for practical federated vision-language adaptation scenarios where client data distributions are inherently heterogeneous.

\begin{table*}[t]
\centering
\caption{Performance on CLIP datasets under IID and non-IID partitions (Part 1).}
\label{tab:clip-full-part1}

\footnotesize
\setlength{\tabcolsep}{2.0pt}
\renewcommand{\arraystretch}{1.0}

\begin{adjustbox}{max width=\textwidth}
\begin{tabular}{@{}llcccccccccc@{}}

\toprule

\multirow{2}{*}{TTA}
& \multirow{2}{*}{Fed.}
& \multicolumn{2}{c}{Flower102}
& \multicolumn{2}{c}{Caltech101}
& \multicolumn{2}{c}{OxfordPets}
& \multicolumn{2}{c}{DTD}
& \multicolumn{2}{c}{Food101} \\

\cmidrule(l{1pt}r{1pt}){3-4}
\cmidrule(l{1pt}r{1pt}){5-6}
\cmidrule(l{1pt}r{1pt}){7-8}
\cmidrule(l{1pt}r{1pt}){9-10}
\cmidrule(l{1pt}r{1pt}){11-12}

& & IID & Non-IID & IID & Non-IID & IID & Non-IID & IID & Non-IID & IID & Non-IID \\

\midrule

\multirow{9}{*}{TPT}
& None-Fed & 14.08\,$\pm$\,0.22 & 22.26\,$\pm$\,0.20 & 26.07\,$\pm$\,0.26 & 33.25\,$\pm$\,0.33 & 21.09\,$\pm$\,0.10 & 28.23\,$\pm$\,0.34 & 10.09\,$\pm$\,0.11 & 11.07\,$\pm$\,0.16 & 18.13\,$\pm$\,0.17 & 25.07\,$\pm$\,0.26 \\
& FedAvg & 27.78\,$\pm$\,1.29 & 31.13\,$\pm$\,1.17 & 37.80\,$\pm$\,1.31 & 42.13\,$\pm$\,1.20 & 34.79\,$\pm$\,1.25 & 38.00\,$\pm$\,1.41 & 20.53\,$\pm$\,1.25 & 22.50\,$\pm$\,1.29 & 31.82\,$\pm$\,1.29 & 38.42\,$\pm$\,1.40 \\
& FedAvgM & 28.06\,$\pm$\,1.26 & 31.96\,$\pm$\,1.15 & 38.37\,$\pm$\,1.26 & 42.54\,$\pm$\,1.12 & 34.95\,$\pm$\,1.23 & 38.57\,$\pm$\,1.37 & 21.30\,$\pm$\,1.16 & 22.88\,$\pm$\,1.27 & 32.62\,$\pm$\,1.24 & \underline{38.68\,$\pm$\,1.37} \\
& FedProx & \textbf{29.74\,$\pm$\,1.25} & 34.52\,$\pm$\,1.16 & \textbf{41.78\,$\pm$\,1.14} & 45.56\,$\pm$\,1.05 & \underline{35.87\,$\pm$\,1.06} & 40.53\,$\pm$\,0.19 & 21.47\,$\pm$\,0.25 & \underline{23.44\,$\pm$\,0.06} & \textbf{33.64\,$\pm$\,0.36} & \textbf{38.82\,$\pm$\,0.12} \\
& FedAMP & 26.55\,$\pm$\,1.32 & 38.76\,$\pm$\,0.16 & 38.53\,$\pm$\,0.99 & 50.55\,$\pm$\,1.35 & 33.44\,$\pm$\,0.34 & 44.69\,$\pm$\,1.27 & 20.95\,$\pm$\,1.07 & 21.91\,$\pm$\,1.13 & 31.42\,$\pm$\,0.80 & 36.45\,$\pm$\,1.27 \\
& pFedGraph & 27.44\,$\pm$\,1.27 & \underline{40.72\,$\pm$\,1.05} & 37.40\,$\pm$\,1.39 & \underline{52.72\,$\pm$\,1.38} & 34.51\,$\pm$\,1.16 & 45.66\,$\pm$\,1.15 & 22.29\,$\pm$\,1.33 & 23.38\,$\pm$\,1.14 & 31.54\,$\pm$\,0.99 & 36.50\,$\pm$\,2.17 \\
& FedCTTA & 28.78\,$\pm$\,0.61 & 39.92\,$\pm$\,1.13 & 36.54\,$\pm$\,0.66 & 51.96\,$\pm$\,0.89 & 33.56\,$\pm$\,0.97 & \underline{47.51\,$\pm$\,1.17} & \underline{22.69\,$\pm$\,0.81} & 22.58\,$\pm$\,0.92 & 32.39\,$\pm$\,0.91 & 35.81\,$\pm$\,1.11 \\
& pFedMoAP & 24.65\,$\pm$\,1.43 & 37.93\,$\pm$\,1.51 & 35.50\,$\pm$\,1.40 & 50.63\,$\pm$\,1.54 & 33.03\,$\pm$\,1.31 & 44.18\,$\pm$\,1.29 & 18.06\,$\pm$\,1.29 & 20.75\,$\pm$\,1.20 & 29.58\,$\pm$\,1.23 & 32.98\,$\pm$\,1.47 \\
& FedRAPP & \underline{29.52\,$\pm$\,0.55} & \textbf{42.79\,$\pm$\,0.99} & \underline{39.51\,$\pm$\,0.62} & \textbf{54.76\,$\pm$\,0.88} & \textbf{36.54\,$\pm$\,0.91} & \textbf{48.79\,$\pm$\,1.13} & \textbf{22.76\,$\pm$\,0.73} & \textbf{24.76\,$\pm$\,0.87} & \underline{33.52\,$\pm$\,0.90} & 37.75\,$\pm$\,1.03 \\

\midrule

\multirow{9}{*}{DART}
& None-Fed & \textbf{68.42\,$\pm$\,0.22} & \textbf{68.66\,$\pm$\,0.18} & \textbf{94.01\,$\pm$\,0.15} & \textbf{94.42\,$\pm$\,0.04} & \textbf{86.78\,$\pm$\,0.15} & \textbf{87.49\,$\pm$\,0.21} & \textbf{45.14\,$\pm$\,0.19} & \textbf{45.26\,$\pm$\,0.08} & \textbf{82.07\,$\pm$\,0.16} & \textbf{82.71\,$\pm$\,0.08} \\
& FedAvg & 20.54\,$\pm$\,2.21 & 24.38\,$\pm$\,1.30 & 32.63\,$\pm$\,2.11 & 36.48\,$\pm$\,2.05 & 26.50\,$\pm$\,1.29 & 31.36\,$\pm$\,0.90 & 19.03\,$\pm$\,1.14 & 19.22\,$\pm$\,3.29 & 41.08\,$\pm$\,2.36 & 41.35\,$\pm$\,1.09 \\
& FedAvgM & 20.66\,$\pm$\,2.20 & 24.99\,$\pm$\,1.27 & 32.71\,$\pm$\,1.99 & 37.00\,$\pm$\,1.92 & 26.54\,$\pm$\,1.25 & 31.94\,$\pm$\,0.91 & 19.27\,$\pm$\,1.12 & 19.72\,$\pm$\,3.01 & 41.28\,$\pm$\,2.31 & 42.02\,$\pm$\,1.01 \\
& FedProx & 19.59\,$\pm$\,1.11 & 22.59\,$\pm$\,1.15 & 31.46\,$\pm$\,0.68 & 32.60\,$\pm$\,1.18 & 26.45\,$\pm$\,0.91 & 29.47\,$\pm$\,2.42 & 20.04\,$\pm$\,1.16 & 19.80\,$\pm$\,1.38 & 40.94\,$\pm$\,2.17 & 41.27\,$\pm$\,2.40 \\
& FedAMP & \underline{44.03\,$\pm$\,1.20} & 42.91\,$\pm$\,1.31 & \underline{54.88\,$\pm$\,1.40} & 55.00\,$\pm$\,1.19 & \underline{48.88\,$\pm$\,1.38} & 49.91\,$\pm$\,1.34 & 25.86\,$\pm$\,1.27 & 25.75\,$\pm$\,1.31 & 45.66\,$\pm$\,1.13 & 45.70\,$\pm$\,1.09 \\
& pFedGraph & 42.10\,$\pm$\,1.10 & 42.92\,$\pm$\,1.39 & 52.11\,$\pm$\,1.02 & 53.02\,$\pm$\,2.22 & 47.04\,$\pm$\,1.17 & 48.11\,$\pm$\,1.10 & 24.46\,$\pm$\,0.88 & 24.46\,$\pm$\,1.28 & 44.00\,$\pm$\,1.25 & 44.71\,$\pm$\,1.05 \\
& FedCTTA & 42.46\,$\pm$\,1.01 & 42.64\,$\pm$\,1.11 & 52.90\,$\pm$\,1.07 & 54.41\,$\pm$\,1.10 & 46.46\,$\pm$\,0.93 & 48.16\,$\pm$\,1.21 & 24.29\,$\pm$\,0.87 & 25.18\,$\pm$\,0.81 & 42.79\,$\pm$\,0.71 & 45.42\,$\pm$\,0.97 \\
& pFedMoAP & 38.60\,$\pm$\,1.33 & 38.92\,$\pm$\,1.51 & 49.88\,$\pm$\,1.50 & 50.78\,$\pm$\,1.55 & 43.58\,$\pm$\,1.12 & 45.94\,$\pm$\,1.16 & 21.90\,$\pm$\,1.23 & 22.25\,$\pm$\,1.30 & 41.39\,$\pm$\,1.27 & 42.78\,$\pm$\,1.34 \\
& FedRAPP & 43.10\,$\pm$\,0.83 & \underline{43.11\,$\pm$\,1.06} & 54.10\,$\pm$\,0.94 & \underline{55.06\,$\pm$\,1.02} & 48.14\,$\pm$\,0.88 & \underline{50.00\,$\pm$\,1.07} & \underline{26.02\,$\pm$\,0.76} & \underline{26.19\,$\pm$\,0.77} & \underline{45.68\,$\pm$\,0.55} & \underline{46.47\,$\pm$\,0.72} \\

\midrule

\multirow{9}{*}{SWAPPROMPT}
& None-Fed & 67.51\,$\pm$\,0.13 & 68.61\,$\pm$\,0.16 & 93.20\,$\pm$\,0.19 & \underline{93.62\,$\pm$\,0.05} & 88.10\,$\pm$\,0.14 & 88.82\,$\pm$\,0.10 & 44.55\,$\pm$\,0.13 & \underline{45.27\,$\pm$\,0.08} & 83.62\,$\pm$\,0.10 & 83.94\,$\pm$\,0.13 \\
& FedAvg & 67.54\,$\pm$\,0.17 & 67.99\,$\pm$\,0.18 & 93.38\,$\pm$\,0.26 & 93.41\,$\pm$\,0.25 & 88.32\,$\pm$\,0.17 & 88.47\,$\pm$\,0.21 & 44.40\,$\pm$\,0.09 & 44.64\,$\pm$\,0.25 & 84.18\,$\pm$\,0.31 & 82.88\,$\pm$\,0.37 \\
& FedAvgM & \underline{68.01\,$\pm$\,0.15} & 68.48\,$\pm$\,0.17 & \underline{94.01\,$\pm$\,0.23} & 93.29\,$\pm$\,0.19 & \underline{88.56\,$\pm$\,0.16} & \underline{88.88\,$\pm$\,0.18} & 44.90\,$\pm$\,0.08 & 44.98\,$\pm$\,0.19 & 83.91\,$\pm$\,0.26 & \underline{85.26\,$\pm$\,0.25} \\
& FedProx & 67.78\,$\pm$\,0.18 & 67.96\,$\pm$\,0.26 & 93.45\,$\pm$\,0.12 & 93.56\,$\pm$\,0.04 & 88.23\,$\pm$\,0.16 & 88.41\,$\pm$\,0.36 & 44.86\,$\pm$\,0.16 & 44.75\,$\pm$\,0.12 & \underline{84.26\,$\pm$\,0.28} & 83.41\,$\pm$\,0.10 \\
& FedAMP & 67.58\,$\pm$\,0.14 & \underline{68.69\,$\pm$\,0.10} & 93.36\,$\pm$\,0.22 & 93.32\,$\pm$\,0.23 & 88.38\,$\pm$\,0.31 & 88.83\,$\pm$\,0.15 & 43.95\,$\pm$\,0.29 & 45.01\,$\pm$\,0.36 & 84.11\,$\pm$\,0.27 & 84.19\,$\pm$\,0.24 \\
& pFedGraph & 67.76\,$\pm$\,0.38 & 68.21\,$\pm$\,0.14 & 93.41\,$\pm$\,0.34 & 93.57\,$\pm$\,0.23 & 87.87\,$\pm$\,0.39 & 88.53\,$\pm$\,0.08 & \textbf{44.99\,$\pm$\,0.17} & 44.74\,$\pm$\,0.10 & 84.21\,$\pm$\,0.11 & 84.18\,$\pm$\,0.35 \\
& FedCTTA & 67.10\,$\pm$\,0.15 & 68.13\,$\pm$\,0.16 & 91.98\,$\pm$\,0.12 & 93.03\,$\pm$\,0.17 & 88.50\,$\pm$\,0.11 & 88.47\,$\pm$\,0.14 & 43.98\,$\pm$\,0.12 & 44.26\,$\pm$\,0.04 & 83.37\,$\pm$\,0.09 & 84.05\,$\pm$\,0.10 \\
& pFedMoAP & 64.67\,$\pm$\,0.41 & 66.02\,$\pm$\,0.44 & 89.67\,$\pm$\,0.37 & 89.23\,$\pm$\,0.32 & 84.08\,$\pm$\,0.51 & 85.30\,$\pm$\,0.57 & 40.26\,$\pm$\,0.39 & 41.27\,$\pm$\,0.51 & 79.93\,$\pm$\,0.61 & 80.84\,$\pm$\,0.65 \\
& FedRAPP & \textbf{68.95\,$\pm$\,0.12} & \textbf{69.19\,$\pm$\,0.09} & \textbf{94.56\,$\pm$\,0.13} & \textbf{93.83\,$\pm$\,0.13} & \textbf{89.01\,$\pm$\,0.14} & \textbf{89.06\,$\pm$\,0.20} & \underline{44.96\,$\pm$\,0.11} & \textbf{45.61\,$\pm$\,0.03} & \textbf{84.63\,$\pm$\,0.13} & \textbf{85.81\,$\pm$\,0.14} \\

\midrule

\multirow{9}{*}{ADAPROMPT}
& None-Fed & \underline{71.87\,$\pm$\,0.09} & \underline{73.11\,$\pm$\,0.13} & 93.88\,$\pm$\,0.13 & 94.02\,$\pm$\,0.04 & 88.04\,$\pm$\,0.11 & 86.11\,$\pm$\,0.08 & 44.94\,$\pm$\,0.09 & 44.40\,$\pm$\,0.11 & 84.29\,$\pm$\,0.13 & \underline{82.35\,$\pm$\,0.07} \\
& FedAvg & 71.46\,$\pm$\,0.22 & 72.01\,$\pm$\,0.19 & 94.10\,$\pm$\,0.15 & 94.00\,$\pm$\,0.22 & 89.86\,$\pm$\,0.33 & \underline{87.22\,$\pm$\,0.27} & 44.91\,$\pm$\,0.24 & \underline{44.56\,$\pm$\,0.08} & \textbf{84.66\,$\pm$\,0.32} & 77.54\,$\pm$\,0.13 \\
& FedAvgM & 70.24\,$\pm$\,0.23 & 70.61\,$\pm$\,0.18 & 93.65\,$\pm$\,0.12 & \underline{95.26\,$\pm$\,0.21} & 89.92\,$\pm$\,0.27 & 86.58\,$\pm$\,0.25 & 44.92\,$\pm$\,0.20 & 44.03\,$\pm$\,0.09 & 84.00\,$\pm$\,0.29 & 81.75\,$\pm$\,0.12 \\
& FedProx & 71.43\,$\pm$\,0.33 & 71.49\,$\pm$\,0.10 & \underline{94.27\,$\pm$\,0.24} & 93.83\,$\pm$\,0.33 & 90.26\,$\pm$\,0.30 & 87.09\,$\pm$\,0.15 & 44.75\,$\pm$\,0.23 & 44.30\,$\pm$\,0.25 & 83.26\,$\pm$\,0.42 & 77.51\,$\pm$\,0.06 \\
& FedAMP & 69.66\,$\pm$\,0.17 & 72.30\,$\pm$\,0.38 & 93.31\,$\pm$\,0.30 & 94.95\,$\pm$\,0.18 & 87.39\,$\pm$\,0.09 & 85.29\,$\pm$\,0.13 & 44.54\,$\pm$\,0.16 & 44.41\,$\pm$\,0.13 & 82.89\,$\pm$\,0.06 & 82.11\,$\pm$\,0.37 \\
& pFedGraph & 71.56\,$\pm$\,0.18 & 72.07\,$\pm$\,0.36 & 93.97\,$\pm$\,0.21 & 93.65\,$\pm$\,0.04 & \textbf{90.32\,$\pm$\,0.07} & 86.75\,$\pm$\,0.18 & \underline{45.05\,$\pm$\,0.13} & 44.51\,$\pm$\,0.22 & 82.68\,$\pm$\,0.13 & 80.23\,$\pm$\,0.36 \\
& FedCTTA & 69.43\,$\pm$\,0.13 & 72.12\,$\pm$\,0.16 & 91.56\,$\pm$\,0.19 & 94.67\,$\pm$\,0.18 & 87.84\,$\pm$\,0.10 & 85.01\,$\pm$\,0.12 & 44.68\,$\pm$\,0.14 & 44.11\,$\pm$\,0.12 & 81.79\,$\pm$\,0.15 & 82.00\,$\pm$\,0.14 \\
& pFedMoAP & 67.10\,$\pm$\,0.22 & 68.37\,$\pm$\,0.37 & 89.25\,$\pm$\,0.26 & 91.74\,$\pm$\,0.19 & 84.91\,$\pm$\,0.15 & 85.39\,$\pm$\,0.21 & 41.77\,$\pm$\,0.17 & 41.57\,$\pm$\,0.28 & 79.21\,$\pm$\,0.30 & 79.75\,$\pm$\,0.41 \\
& FedRAPP & \textbf{71.95\,$\pm$\,0.11} & \textbf{73.79\,$\pm$\,0.07} & \textbf{94.31\,$\pm$\,0.09} & \textbf{95.92\,$\pm$\,0.13} & \underline{90.29\,$\pm$\,0.06} & \textbf{89.18\,$\pm$\,0.05} & \textbf{45.29\,$\pm$\,0.11} & \textbf{45.11\,$\pm$\,0.11} & \underline{84.50\,$\pm$\,0.09} & \textbf{82.44\,$\pm$\,0.13} \\

\bottomrule

\end{tabular}
\end{adjustbox}
\end{table*}

\begin{table*}[t]
\centering
\caption{Performance on CLIP datasets under IID and non-IID partitions (Part 2).}
\label{tab:clip-full-part2}

\footnotesize
\setlength{\tabcolsep}{2.0pt}
\renewcommand{\arraystretch}{1.0}

\begin{adjustbox}{max width=\textwidth}
\begin{tabular}{@{}llcccccccc@{}}

\toprule

\multirow{2}{*}{TTA}
& \multirow{2}{*}{Fed.}
& \multicolumn{2}{c}{UCF101}
& \multicolumn{2}{c}{SUN397}
& \multicolumn{2}{c}{FGVCAircraft}
& \multicolumn{2}{c}{StanfordCars} \\

\cmidrule(l{1pt}r{1pt}){3-4}
\cmidrule(l{1pt}r{1pt}){5-6}
\cmidrule(l{1pt}r{1pt}){7-8}
\cmidrule(l{1pt}r{1pt}){9-10}

& & IID & Non-IID & IID & Non-IID & IID & Non-IID & IID & Non-IID \\

\midrule

\multirow{9}{*}{TPT}
& None-Fed & 15.07\,$\pm$\,0.27 & 17.30\,$\pm$\,0.22 & 9.09\,$\pm$\,0.12 & 11.08\,$\pm$\,0.21 & 7.08\,$\pm$\,0.19 & 10.08\,$\pm$\,0.16 & 13.10\,$\pm$\,0.12 & 14.05\,$\pm$\,0.23 \\
& FedAvg & 30.63\,$\pm$\,1.20 & 32.64\,$\pm$\,1.09 & 19.51\,$\pm$\,2.24 & 21.60\,$\pm$\,2.08 & \underline{14.21\,$\pm$\,1.38} & \underline{17.06\,$\pm$\,0.93} & 28.62\,$\pm$\,1.50 & 29.63\,$\pm$\,1.61 \\
& FedAvgM & \textbf{31.54\,$\pm$\,1.19} & \underline{33.58\,$\pm$\,1.08} & 20.12\,$\pm$\,2.19 & 22.53\,$\pm$\,2.07 & \textbf{14.42\,$\pm$\,1.21} & \textbf{17.97\,$\pm$\,0.81} & \textbf{29.53\,$\pm$\,1.43} & \textbf{30.20\,$\pm$\,1.57} \\
& FedProx & \underline{30.80\,$\pm$\,1.11} & \textbf{33.83\,$\pm$\,1.06} & 20.35\,$\pm$\,1.17 & 22.37\,$\pm$\,1.12 & 12.90\,$\pm$\,0.82 & 15.98\,$\pm$\,1.03 & \underline{28.77\,$\pm$\,1.76} & \underline{29.75\,$\pm$\,0.72} \\
& FedAMP & 27.51\,$\pm$\,1.21 & 30.39\,$\pm$\,1.39 & 19.88\,$\pm$\,1.31 & 21.93\,$\pm$\,0.67 & 13.43\,$\pm$\,1.29 & 16.41\,$\pm$\,0.72 & 25.37\,$\pm$\,0.59 & 26.46\,$\pm$\,1.27 \\
& pFedGraph & 28.58\,$\pm$\,1.10 & 31.46\,$\pm$\,1.25 & \underline{21.38\,$\pm$\,1.17} & \underline{23.46\,$\pm$\,1.06} & 12.60\,$\pm$\,1.12 & 15.49\,$\pm$\,1.31 & 26.49\,$\pm$\,1.27 & 27.56\,$\pm$\,0.74 \\
& FedCTTA & 27.09\,$\pm$\,0.81 & 29.47\,$\pm$\,0.81 & 18.92\,$\pm$\,0.75 & 22.70\,$\pm$\,0.53 & 10.76\,$\pm$\,0.51 & 14.13\,$\pm$\,0.41 & 25.45\,$\pm$\,0.37 & 25.79\,$\pm$\,0.51 \\
& pFedMoAP & 24.18\,$\pm$\,1.29 & 27.37\,$\pm$\,1.31 & 18.52\,$\pm$\,1.39 & 20.63\,$\pm$\,1.23 & 8.80\,$\pm$\,1.29 & 13.20\,$\pm$\,1.30 & 22.37\,$\pm$\,1.36 & 24.72\,$\pm$\,1.01 \\
& FedRAPP & 28.78\,$\pm$\,0.65 & 31.78\,$\pm$\,0.73 & \textbf{21.81\,$\pm$\,0.71} & \textbf{23.81\,$\pm$\,0.50} & 13.61\,$\pm$\,0.47 & 16.58\,$\pm$\,0.37 & 26.76\,$\pm$\,0.33 & 27.76\,$\pm$\,0.47 \\

\midrule

\multirow{9}{*}{DART}
& None-Fed & \textbf{66.21\,$\pm$\,1.11} & \textbf{66.99\,$\pm$\,1.08} & \textbf{60.25\,$\pm$\,1.09} & \textbf{60.18\,$\pm$\,2.11} & \textbf{23.13\,$\pm$\,0.99} & \textbf{23.43\,$\pm$\,0.43} & \textbf{65.19\,$\pm$\,0.65} & \textbf{65.18\,$\pm$\,1.11} \\
& FedAvg & 31.32\,$\pm$\,2.06 & 30.24\,$\pm$\,2.23 & 25.18\,$\pm$\,2.30 & 26.75\,$\pm$\,2.32 & 12.03\,$\pm$\,2.16 & 13.23\,$\pm$\,0.99 & 26.53\,$\pm$\,1.39 & 27.08\,$\pm$\,2.65 \\
& FedAvgM & 31.51\,$\pm$\,1.93 & 30.91\,$\pm$\,2.21 & 25.24\,$\pm$\,2.28 & 27.28\,$\pm$\,2.26 & 12.10\,$\pm$\,2.11 & 13.75\,$\pm$\,0.92 & 26.54\,$\pm$\,1.36 & 27.77\,$\pm$\,2.66 \\
& FedProx & 32.32\,$\pm$\,1.07 & 32.47\,$\pm$\,2.34 & 24.31\,$\pm$\,1.60 & 26.00\,$\pm$\,2.01 & 13.61\,$\pm$\,1.25 & 14.01\,$\pm$\,2.25 & 27.67\,$\pm$\,2.02 & 26.23\,$\pm$\,1.67 \\
& FedAMP & 35.23\,$\pm$\,1.19 & 36.03\,$\pm$\,1.18 & 25.89\,$\pm$\,1.41 & 27.82\,$\pm$\,1.36 & 14.83\,$\pm$\,1.24 & 14.65\,$\pm$\,1.39 & 29.00\,$\pm$\,0.95 & 29.61\,$\pm$\,1.40 \\
& pFedGraph & \underline{36.00\,$\pm$\,1.28} & 35.34\,$\pm$\,0.73 & 25.02\,$\pm$\,1.30 & 25.80\,$\pm$\,1.35 & 13.86\,$\pm$\,1.68 & 14.54\,$\pm$\,1.30 & 27.42\,$\pm$\,1.25 & 27.93\,$\pm$\,1.38 \\
& FedCTTA & 34.53\,$\pm$\,0.55 & 35.08\,$\pm$\,0.60 & 23.67\,$\pm$\,0.83 & 26.37\,$\pm$\,0.92 & 15.10\,$\pm$\,1.07 & 13.64\,$\pm$\,0.73 & 27.04\,$\pm$\,0.71 & 29.29\,$\pm$\,0.65 \\
& pFedMoAP & 30.96\,$\pm$\,1.68 & 31.59\,$\pm$\,1.53 & 22.09\,$\pm$\,1.71 & 23.87\,$\pm$\,1.43 & 11.59\,$\pm$\,1.81 & 12.48\,$\pm$\,1.74 & 24.49\,$\pm$\,1.30 & 25.16\,$\pm$\,1.61 \\
& FedRAPP & 35.61\,$\pm$\,0.57 & \underline{36.15\,$\pm$\,0.38} & \underline{26.02\,$\pm$\,0.69} & \underline{28.10\,$\pm$\,0.75} & \underline{16.22\,$\pm$\,1.03} & \underline{16.63\,$\pm$\,0.45} & \underline{29.15\,$\pm$\,0.57} & \underline{30.09\,$\pm$\,0.34} \\

\midrule

\multirow{9}{*}{SWAPPROMPT}
& None-Fed & 65.20\,$\pm$\,0.18 & 65.67\,$\pm$\,0.10 & 62.63\,$\pm$\,0.16 & 62.69\,$\pm$\,0.14 & 23.53\,$\pm$\,0.12 & 23.56\,$\pm$\,0.08 & \underline{65.23\,$\pm$\,0.12} & 65.21\,$\pm$\,0.06 \\
& FedAvg & 65.89\,$\pm$\,0.12 & 64.22\,$\pm$\,0.31 & 63.04\,$\pm$\,0.18 & 61.58\,$\pm$\,0.26 & \underline{23.73\,$\pm$\,0.12} & 22.83\,$\pm$\,0.22 & 65.10\,$\pm$\,0.23 & 64.97\,$\pm$\,0.21 \\
& FedAvgM & 65.52\,$\pm$\,0.14 & 65.81\,$\pm$\,0.18 & 62.82\,$\pm$\,0.15 & 63.44\,$\pm$\,0.21 & 23.15\,$\pm$\,0.10 & 23.36\,$\pm$\,0.17 & 65.04\,$\pm$\,0.24 & 64.98\,$\pm$\,0.18 \\
& FedProx & \textbf{66.15\,$\pm$\,0.23} & 65.42\,$\pm$\,0.30 & 63.16\,$\pm$\,0.17 & 62.50\,$\pm$\,0.05 & 23.69\,$\pm$\,0.25 & 23.72\,$\pm$\,0.21 & 65.18\,$\pm$\,0.22 & 64.95\,$\pm$\,0.16 \\
& FedAMP & 65.72\,$\pm$\,0.17 & \underline{66.06\,$\pm$\,0.15} & 63.21\,$\pm$\,0.06 & 63.23\,$\pm$\,0.15 & 23.16\,$\pm$\,0.21 & \underline{23.97\,$\pm$\,0.24} & 64.96\,$\pm$\,0.40 & \textbf{65.61\,$\pm$\,0.17} \\
& pFedGraph & 65.97\,$\pm$\,0.27 & 65.94\,$\pm$\,0.25 & \underline{63.22\,$\pm$\,0.36} & \underline{63.46\,$\pm$\,0.10} & 23.63\,$\pm$\,0.22 & 22.84\,$\pm$\,0.29 & 64.97\,$\pm$\,0.06 & 64.89\,$\pm$\,0.24 \\
& FedCTTA & 65.43\,$\pm$\,0.13 & 65.72\,$\pm$\,0.16 & 62.12\,$\pm$\,0.13 & 62.93\,$\pm$\,0.15 & 22.06\,$\pm$\,0.13 & 21.11\,$\pm$\,0.12 & 62.96\,$\pm$\,0.13 & 64.26\,$\pm$\,0.16 \\
& pFedMoAP & 61.53\,$\pm$\,0.46 & 61.54\,$\pm$\,0.39 & 60.26\,$\pm$\,0.27 & 59.08\,$\pm$\,0.34 & 20.67\,$\pm$\,0.31 & 20.86\,$\pm$\,0.35 & 60.80\,$\pm$\,0.73 & 61.22\,$\pm$\,0.61 \\
& FedRAPP & \underline{66.06\,$\pm$\,0.08} & \textbf{66.54\,$\pm$\,0.09} & \textbf{63.32\,$\pm$\,0.11} & \textbf{63.96\,$\pm$\,0.05} & \textbf{23.88\,$\pm$\,0.14} & \textbf{24.03\,$\pm$\,0.10} & \textbf{65.55\,$\pm$\,0.10} & \underline{65.56\,$\pm$\,0.13} \\

\midrule

\multirow{9}{*}{ADAPROMPT}
& None-Fed & 66.12\,$\pm$\,0.12 & 66.29\,$\pm$\,0.09 & \underline{65.37\,$\pm$\,0.11} & 65.14\,$\pm$\,0.10 & 22.98\,$\pm$\,0.19 & 23.16\,$\pm$\,0.13 & 63.96\,$\pm$\,0.13 & 64.01\,$\pm$\,0.08 \\
& FedAvg & \underline{66.99\,$\pm$\,0.38} & 65.02\,$\pm$\,0.10 & 63.90\,$\pm$\,0.36 & 60.19\,$\pm$\,0.41 & \underline{23.03\,$\pm$\,0.18} & 22.88\,$\pm$\,0.09 & 64.01\,$\pm$\,0.09 & 63.48\,$\pm$\,0.33 \\
& FedAvgM & 66.19\,$\pm$\,0.37 & \underline{66.98\,$\pm$\,0.08} & 65.12\,$\pm$\,0.34 & 64.80\,$\pm$\,0.38 & 22.27\,$\pm$\,0.16 & 22.72\,$\pm$\,0.10 & 63.42\,$\pm$\,0.08 & \underline{65.57\,$\pm$\,0.29} \\
& FedProx & \textbf{67.03\,$\pm$\,0.22} & 66.93\,$\pm$\,0.30 & 60.39\,$\pm$\,0.28 & 60.33\,$\pm$\,0.25 & 23.01\,$\pm$\,0.19 & 23.01\,$\pm$\,0.12 & 63.99\,$\pm$\,0.23 & 63.64\,$\pm$\,0.16 \\
& FedAMP & 65.39\,$\pm$\,0.07 & 66.96\,$\pm$\,0.12 & 62.28\,$\pm$\,0.29 & \underline{65.18\,$\pm$\,0.14} & \textbf{23.17\,$\pm$\,0.09} & 22.99\,$\pm$\,0.31 & 58.65\,$\pm$\,0.11 & 63.96\,$\pm$\,0.42 \\
& pFedGraph & 66.91\,$\pm$\,0.24 & 64.59\,$\pm$\,0.12 & 59.99\,$\pm$\,0.37 & 60.69\,$\pm$\,0.32 & 22.92\,$\pm$\,0.15 & \underline{23.21\,$\pm$\,0.06} & \underline{64.07\,$\pm$\,0.22} & 63.61\,$\pm$\,0.37 \\
& FedCTTA & 64.66\,$\pm$\,0.13 & 64.82\,$\pm$\,0.12 & 64.04\,$\pm$\,0.19 & 62.56\,$\pm$\,0.21 & 21.08\,$\pm$\,0.22 & 20.94\,$\pm$\,0.27 & 63.12\,$\pm$\,0.31 & 64.93\,$\pm$\,0.20 \\
& pFedMoAP & 62.49\,$\pm$\,0.41 & 62.82\,$\pm$\,0.38 & 61.61\,$\pm$\,0.32 & 61.72\,$\pm$\,0.30 & 18.65\,$\pm$\,0.21 & 19.67\,$\pm$\,0.27 & 61.05\,$\pm$\,0.30 & 62.55\,$\pm$\,0.28 \\
& FedRAPP & 66.91\,$\pm$\,0.06 & \textbf{67.73\,$\pm$\,0.12} & \textbf{65.79\,$\pm$\,0.07} & \textbf{65.52\,$\pm$\,0.10} & 22.79\,$\pm$\,0.14 & \textbf{23.25\,$\pm$\,0.11} & \textbf{64.15\,$\pm$\,0.04} & \textbf{66.17\,$\pm$\,0.09} \\

\bottomrule

\end{tabular}
\end{adjustbox}
\end{table*}

\begin{table*}[t]
\centering
\caption{Performance on ImageNet variants under IID and non-IID partitions with
TPT and DART local adaptation, reported as mean\,$\pm$\,standard deviation.
This table complements Table~\ref{tab:imagenet-variants} of the main paper,
which covers SwapPrompt and AdaPrompt. The \textbf{best} and
\underline{second-best} results within each block are highlighted in bold and
underlined, respectively.}
\label{tab:imagenet-tpt-dart}
\footnotesize
\setlength{\tabcolsep}{2.6pt}
\renewcommand{\arraystretch}{1.03}
\begin{adjustbox}{max width=\textwidth}
\begin{tabular}{llcccccccccccc}
\toprule
\multirow{2}{*}{TTA}
& \multirow{2}{*}{Fed.}
& \multicolumn{2}{c}{ImageNet}
& \multicolumn{2}{c}{ImageNet-V2}
& \multicolumn{2}{c}{ImageNet-A}
& \multicolumn{2}{c}{ImageNet-S}
& \multicolumn{2}{c}{ImageNet-R}
& \multicolumn{2}{c}{Average} \\
\cmidrule(lr){3-4}
\cmidrule(lr){5-6}
\cmidrule(lr){7-8}
\cmidrule(lr){9-10}
\cmidrule(lr){11-12}
\cmidrule(lr){13-14}
& & IID & Non-IID & IID & Non-IID & IID & Non-IID & IID & Non-IID & IID & Non-IID & IID & Non-IID \\
\midrule
\multirow{9}{*}{TPT}
& None-Fed  & 12.12\,$\pm$\,0.13 & 12.41\,$\pm$\,0.06 & 10.78\,$\pm$\,0.12 & 10.88\,$\pm$\,0.15 & 6.97\,$\pm$\,0.08 & 7.35\,$\pm$\,0.08 & 8.02\,$\pm$\,0.14 & 8.35\,$\pm$\,0.04 & 9.82\,$\pm$\,0.09 & 9.71\,$\pm$\,0.12 & 9.54 & 9.74 \\
& FedAvg    & 14.91\,$\pm$\,0.41 & 14.54\,$\pm$\,0.13 & 13.34\,$\pm$\,0.32 & 13.30\,$\pm$\,0.14 & 15.37\,$\pm$\,0.25 & 15.60\,$\pm$\,0.05 & 17.86\,$\pm$\,0.37 & 18.33\,$\pm$\,0.12 & 19.60\,$\pm$\,0.32 & 19.81\,$\pm$\,0.24 & 16.22 & 16.32 \\
& FedAvgM  & 16.40\,$\pm$\,0.39 & 15.59\,$\pm$\,0.14 & 14.37\,$\pm$\,0.26 & 14.57\,$\pm$\,0.13 & 16.63\,$\pm$\,0.19 & 16.89\,$\pm$\,0.07 & 19.12\,$\pm$\,0.41 & 19.37\,$\pm$\,0.15 & 21.14\,$\pm$\,0.31 & 20.93\,$\pm$\,0.27 & 17.53 & 17.47 \\
& FedProx   & 15.93\,$\pm$\,0.39 & 15.48\,$\pm$\,0.21 & 14.27\,$\pm$\,0.41 & 14.18\,$\pm$\,0.35 & 16.36\,$\pm$\,0.22 & 16.60\,$\pm$\,0.06 & 18.96\,$\pm$\,0.20 & 19.17\,$\pm$\,0.39 & 20.76\,$\pm$\,0.04 & 20.84\,$\pm$\,0.20 & 17.26 & 17.25 \\
& FedAMP    & 28.12\,$\pm$\,0.08 & 25.53\,$\pm$\,0.17 & \textbf{28.33\,$\pm$\,0.26} & \underline{26.19\,$\pm$\,0.30} & \textbf{29.32\,$\pm$\,0.35} & \underline{29.49\,$\pm$\,0.26} & 27.90\,$\pm$\,0.30 & 29.29\,$\pm$\,0.11 & 30.57\,$\pm$\,0.33 & 30.71\,$\pm$\,0.37 & 28.85 & 28.24 \\
& pFedGraph & 26.98\,$\pm$\,0.23 & 26.47\,$\pm$\,0.22 & 25.35\,$\pm$\,0.24 & 25.33\,$\pm$\,0.09 & 27.27\,$\pm$\,0.41 & 27.41\,$\pm$\,0.34 & 28.02\,$\pm$\,0.12 & 28.36\,$\pm$\,0.24 & 30.63\,$\pm$\,0.27 & 31.74\,$\pm$\,0.35 & 27.65 & 27.86 \\
& FedCTTA   & \underline{28.75\,$\pm$\,0.13} & \underline{27.26\,$\pm$\,0.12} & 24.98\,$\pm$\,0.11 & 26.02\,$\pm$\,0.13 & 25.99\,$\pm$\,0.12 & 29.29\,$\pm$\,0.15 & \underline{31.40\,$\pm$\,0.14} & \underline{30.54\,$\pm$\,0.19} & \underline{34.20\,$\pm$\,0.13} & \underline{32.06\,$\pm$\,0.15} & \underline{29.06} & \underline{29.03} \\
& pFedMoAP  & 24.90\,$\pm$\,0.49 & 21.67\,$\pm$\,0.44 & 20.35\,$\pm$\,0.39 & 21.91\,$\pm$\,0.46 & 21.77\,$\pm$\,0.55 & 24.02\,$\pm$\,0.46 & 26.15\,$\pm$\,0.37 & 26.10\,$\pm$\,0.43 & 29.51\,$\pm$\,0.41 & 27.16\,$\pm$\,0.48 & 24.53 & 24.17 \\
& FedRAPP   & \textbf{30.15\,$\pm$\,0.09} & \textbf{27.58\,$\pm$\,0.02} & \underline{25.46\,$\pm$\,0.06} & \textbf{27.38\,$\pm$\,0.12} & \underline{27.48\,$\pm$\,0.07} & \textbf{29.62\,$\pm$\,0.06} & \textbf{32.05\,$\pm$\,0.03} & \textbf{31.42\,$\pm$\,0.13} & \textbf{34.76\,$\pm$\,0.07} & \textbf{32.94\,$\pm$\,0.13} & \textbf{29.98} & \textbf{29.79} \\
\midrule
\multirow{9}{*}{DART}
& None-Fed  & \textbf{67.94\,$\pm$\,0.05} & \textbf{67.67\,$\pm$\,0.07} & \textbf{64.43\,$\pm$\,0.14} & \textbf{64.56\,$\pm$\,0.09} & \textbf{62.42\,$\pm$\,0.13} & \textbf{62.43\,$\pm$\,0.10} & \textbf{51.68\,$\pm$\,0.06} & \textbf{51.67\,$\pm$\,0.09} & \textbf{81.42\,$\pm$\,0.09} & \textbf{81.53\,$\pm$\,0.09} & \textbf{65.58} & \textbf{65.57} \\
& FedAvg    & 32.21\,$\pm$\,0.29 & 30.11\,$\pm$\,0.26 & 33.12\,$\pm$\,0.24 & 28.95\,$\pm$\,0.33 & 26.26\,$\pm$\,0.05 & 29.03\,$\pm$\,0.21 & 24.25\,$\pm$\,0.04 & 22.11\,$\pm$\,0.06 & 44.18\,$\pm$\,0.19 & 45.03\,$\pm$\,0.42 & 32.00 & 31.05 \\
& FedAvgM  & 32.55\,$\pm$\,0.27 & 30.89\,$\pm$\,0.23 & 33.84\,$\pm$\,0.21 & 29.18\,$\pm$\,0.26 & 26.97\,$\pm$\,0.09 & 29.23\,$\pm$\,0.13 & 24.99\,$\pm$\,0.03 & 22.30\,$\pm$\,0.07 & 44.94\,$\pm$\,0.24 & 45.18\,$\pm$\,0.37 & 32.66 & 31.36 \\
& FedProx   & 33.55\,$\pm$\,0.27 & 33.19\,$\pm$\,0.11 & 32.54\,$\pm$\,0.14 & 33.02\,$\pm$\,0.23 & 25.51\,$\pm$\,0.17 & 32.05\,$\pm$\,0.23 & 25.43\,$\pm$\,0.29 & 25.08\,$\pm$\,0.18 & 46.42\,$\pm$\,0.30 & 46.41\,$\pm$\,0.30 & 32.69 & 33.95 \\
& FedAMP    & 40.00\,$\pm$\,0.21 & 40.91\,$\pm$\,0.35 & 40.81\,$\pm$\,0.43 & 40.87\,$\pm$\,0.29 & 33.97\,$\pm$\,0.07 & \underline{39.90\,$\pm$\,0.21} & 30.93\,$\pm$\,0.20 & \underline{32.84\,$\pm$\,0.38} & 52.91\,$\pm$\,0.24 & 52.90\,$\pm$\,0.22 & 39.72 & \underline{41.48} \\
& pFedGraph & 39.48\,$\pm$\,0.37 & 39.63\,$\pm$\,0.12 & 38.59\,$\pm$\,0.05 & 39.62\,$\pm$\,0.05 & 32.59\,$\pm$\,0.08 & 38.40\,$\pm$\,0.38 & 31.44\,$\pm$\,0.27 & 31.52\,$\pm$\,0.16 & 52.62\,$\pm$\,0.06 & 52.59\,$\pm$\,0.08 & 38.94 & 40.35 \\
& FedCTTA   & 39.48\,$\pm$\,0.10 & 40.54\,$\pm$\,0.11 & 39.44\,$\pm$\,0.12 & 40.90\,$\pm$\,0.13 & 34.17\,$\pm$\,0.09 & 37.91\,$\pm$\,0.15 & 31.18\,$\pm$\,0.12 & 31.80\,$\pm$\,0.13 & 51.14\,$\pm$\,0.10 & 52.12\,$\pm$\,0.11 & 39.08 & 40.65 \\
& pFedMoAP  & 37.03\,$\pm$\,0.44 & 37.97\,$\pm$\,0.43 & 37.07\,$\pm$\,0.48 & 37.71\,$\pm$\,0.35 & 31.76\,$\pm$\,0.47 & 35.48\,$\pm$\,0.42 & 28.16\,$\pm$\,0.39 & 29.06\,$\pm$\,0.35 & 49.71\,$\pm$\,0.40 & 50.00\,$\pm$\,0.36 & 36.75 & 38.04 \\
& FedRAPP   & \underline{40.10\,$\pm$\,0.04} & \underline{41.18\,$\pm$\,0.05} & \underline{41.07\,$\pm$\,0.13} & \underline{41.08\,$\pm$\,0.04} & \underline{35.04\,$\pm$\,0.07} & 39.11\,$\pm$\,0.11 & \underline{32.07\,$\pm$\,0.07} & 32.08\,$\pm$\,0.06 & \underline{53.06\,$\pm$\,0.07} & \underline{53.03\,$\pm$\,0.02} & \underline{40.27} & 41.30 \\
\bottomrule
\end{tabular}
\end{adjustbox}
\end{table*}

\section{Statistical Significance Analysis}
\label{sec:significance}
To assess whether the improvements of FedRAPP are statistically significant, we conduct two non-parametric tests following Dem\v{s}ar~\cite{demvsar2006statistical}, using the mean accuracies of the nine CLIP datasets and five ImageNet variants reported in Tables~\ref{tab:clip-full-part1} and~\ref{tab:clip-full-part2} and in Table~\ref{tab:imagenet-variants} of the main paper, with each dataset treated as one paired sample. We focus on SwapPrompt and AdaPrompt, the two local adaptation methods that remain stable under federation. We first perform a two-sided Wilcoxon signed-rank test between FedRAPP and each of the other federated methods (FedAvg, FedAvgM, FedProx, FedAMP, pFedGraph, FedCTTA and pFedMoAP) and None-Fed, with the Holm correction~\cite{holm1979simple} applied over the eight comparisons. Under the non-IID partition, FedRAPP significantly outperforms every baseline for both adaptation methods (all 16 adjusted $p<0.05$), winning on at least 12 of the 14 datasets against each baseline. Under the IID partition, FedRAPP significantly outperforms None-Fed, FedAvgM, FedAMP, FedCTTA and pFedMoAP for both adaptation methods, while its gains over FedAvg and FedProx (and over pFedGraph with AdaPrompt) are not significant as these methods are already strong when the client data are similar. We further apply the Friedman test, which ranks the nine methods on each dataset and confirms significant differences among them in all four settings ($p<10^{-9}$). FedRAPP achieves the best average rank in every setting, with average ranks of 1.14 and 1.07 for SwapPrompt and AdaPrompt under the non-IID partition, and 2.21 and 2.04 under the IID partition, respectively. The Nemenyi post-hoc test (critical difference of 3.21 at $\alpha=0.05$) further shows that, under the non-IID partition, FedRAPP ranks significantly higher than FedAvg, FedAvgM, FedProx, pFedGraph, FedCTTA, and pFedMoAP for both adaptation methods. These results indicate that, under heterogeneous client distributions, the improvements of FedRAPP are consistent across datasets rather than driven by a few favorable cases, demonstrating the effectiveness of reliability-aware personalized aggregation.

\section{Results with a ResNet-50 Backbone}
\label{sec:rn50}

All results in the main paper use the CLIP ViT-B/16 image encoder. To test whether FedRAPP also works with a different type of encoder, we repeat the experiments on the nine CLIP datasets with the CLIP ResNet-50 (RN50) image encoder, which is a convolutional network rather than a transformer. We keep the federated setup of the main experiments (10 clients, IID and non-IID partitions, three random seeds) and use SwapPrompt and AdaPrompt as the local adaptation methods. Tables~\ref{tab:rn50-part1} and~\ref{tab:rn50-part2} report the results.

All methods score lower with RN50 than with ViT-B/16, because RN50 is a weaker encoder. For example, None-Fed with AdaPrompt reaches a nine-dataset average of 61.28 with RN50 under the IID partition, compared with 66.83 with ViT-B/16. The main findings, however, are the same as with ViT-B/16.

Under the non-IID partition, FedRAPP gives the best results with both adaptation methods. With AdaPrompt, FedRAPP is the best method on all nine datasets. Its nine-dataset average is 62.22, which is 0.76 higher than None-Fed (61.46) and 0.49 higher than the next-best method, FedAMP (61.73). With SwapPrompt, FedRAPP is the best method on eight of the nine datasets. Its average of 61.87 is 1.35 higher than None-Fed (60.52) and 0.67 higher than FedAMP (61.20). The only exception is SUN397, where FedAMP is 0.02 higher (60.99 vs.\ 60.97). With both adaptation methods, FedRAPP is also higher than None-Fed on all nine datasets.

Methods that share a single global prompt lose accuracy under the non-IID partition, as they do with ViT-B/16. With AdaPrompt, the nine-dataset averages of FedAvg (60.24), FedAvgM (59.91) and FedProx (59.36) are all lower than None-Fed (61.46). The largest drop is on OxfordPets, where FedAvg reaches 81.09, compared with 84.30 for None-Fed.

Under the IID partition, the differences between the methods are smaller. FedRAPP still has the highest nine-dataset average with both adaptation methods: 61.46 with SwapPrompt (0.21 higher than FedAvgM) and 62.04 with AdaPrompt (0.22 higher than FedAvgM). It is the best method on eight of the nine datasets with SwapPrompt (all except Caltech101) and on four of the nine with AdaPrompt. On the other five datasets with AdaPrompt, another method is higher by 0.03 to 0.98 points.

pFedMoAP has the lowest accuracy in 34 of the 36 columns, 2.31 to 5.99 points below FedRAPP, which matches its results with ViT-B/16.

Overall, the gains of FedRAPP are not specific to the ViT-B/16 encoder. With the convolutional RN50 encoder, FedRAPP again has the highest nine-dataset average in all four settings (two adaptation methods and two partitions), and its margin over the other federated methods is larger under the non-IID partition (0.49 to 0.67 points) than under the IID partition (0.21 to 0.22 points).

\begin{table*}[t]
\centering
\caption{Performance with the CLIP ResNet-50 (RN50) image encoder on CLIP datasets under IID and non-IID partitions (Part 1). Mean top-1 accuracy\,$\pm$\,standard deviation over three random seeds. \textbf{Best} and \underline{second-best} per column within each block.}
\label{tab:rn50-part1}

\footnotesize
\setlength{\tabcolsep}{2.0pt}
\renewcommand{\arraystretch}{1.0}

\begin{adjustbox}{max width=\textwidth}
\begin{tabular}{@{}llcccccccccc@{}}

\toprule

\multirow{2}{*}{TTA}
& \multirow{2}{*}{Fed.}
& \multicolumn{2}{c}{Flower102}
& \multicolumn{2}{c}{Caltech101}
& \multicolumn{2}{c}{OxfordPets}
& \multicolumn{2}{c}{DTD}
& \multicolumn{2}{c}{Food101} \\

\cmidrule(l{1pt}r{1pt}){3-4}
\cmidrule(l{1pt}r{1pt}){5-6}
\cmidrule(l{1pt}r{1pt}){7-8}
\cmidrule(l{1pt}r{1pt}){9-10}
\cmidrule(l{1pt}r{1pt}){11-12}

& & IID & Non-IID & IID & Non-IID & IID & Non-IID & IID & Non-IID & IID & Non-IID \\

\midrule

\multirow{9}{*}{SWAPPROMPT}
& None-Fed & 62.34\,$\pm$\,0.13 & 63.39\,$\pm$\,0.05 & 85.94\,$\pm$\,0.11 & \underline{86.08\,$\pm$\,0.35} & 83.21\,$\pm$\,0.14 & 83.94\,$\pm$\,0.28 & 41.90\,$\pm$\,0.32 & 42.03\,$\pm$\,0.39 & 74.99\,$\pm$\,0.25 & 75.12\,$\pm$\,0.13 \\
& FedAvg & 62.98\,$\pm$\,0.22 & 62.74\,$\pm$\,0.31 & 86.22\,$\pm$\,0.19 & 85.77\,$\pm$\,0.05 & 83.51\,$\pm$\,0.23 & 83.57\,$\pm$\,0.58 & 41.93\,$\pm$\,0.25 & 41.57\,$\pm$\,0.14 & 75.64\,$\pm$\,0.20 & 74.70\,$\pm$\,0.05 \\
& FedAvgM & 63.08\,$\pm$\,0.61 & 62.15\,$\pm$\,0.30 & 86.50\,$\pm$\,0.20 & 85.46\,$\pm$\,0.04 & \underline{83.85\,$\pm$\,0.30} & 83.58\,$\pm$\,0.26 & 42.27\,$\pm$\,0.24 & 41.27\,$\pm$\,0.06 & \underline{76.47\,$\pm$\,0.22} & 74.10\,$\pm$\,0.10 \\
& FedProx & 63.20\,$\pm$\,0.23 & 61.84\,$\pm$\,0.38 & \underline{86.66\,$\pm$\,0.31} & 85.02\,$\pm$\,0.14 & \underline{83.85\,$\pm$\,0.23} & 83.54\,$\pm$\,0.83 & 42.37\,$\pm$\,0.29 & 40.78\,$\pm$\,0.09 & 75.95\,$\pm$\,0.02 & 73.79\,$\pm$\,0.41 \\
& FedAMP & \underline{63.47\,$\pm$\,0.34} & \underline{64.40\,$\pm$\,0.22} & \textbf{86.99\,$\pm$\,0.10} & 85.99\,$\pm$\,0.41 & 83.70\,$\pm$\,0.11 & \underline{85.28\,$\pm$\,0.21} & \underline{42.39\,$\pm$\,0.07} & \underline{42.29\,$\pm$\,0.84} & 75.78\,$\pm$\,0.59 & 75.10\,$\pm$\,0.71 \\
& pFedGraph & 63.32\,$\pm$\,0.05 & 64.12\,$\pm$\,0.01 & 86.65\,$\pm$\,0.14 & 85.91\,$\pm$\,0.19 & 83.60\,$\pm$\,0.11 & 84.19\,$\pm$\,0.10 & 42.22\,$\pm$\,0.24 & 42.01\,$\pm$\,0.40 & 75.95\,$\pm$\,0.53 & 74.87\,$\pm$\,0.19 \\
& pFedMoAP & 59.43\,$\pm$\,0.74 & 61.13\,$\pm$\,0.75 & 83.44\,$\pm$\,0.04 & 84.46\,$\pm$\,0.04 & 80.08\,$\pm$\,0.34 & 80.13\,$\pm$\,0.85 & 38.68\,$\pm$\,0.07 & 39.75\,$\pm$\,0.87 & 73.10\,$\pm$\,1.06 & 73.44\,$\pm$\,0.61 \\
& FedCTTA & 63.03\,$\pm$\,0.25 & 64.00\,$\pm$\,0.52 & 86.61\,$\pm$\,0.00 & 86.01\,$\pm$\,0.29 & 83.58\,$\pm$\,0.54 & 83.67\,$\pm$\,0.29 & 42.02\,$\pm$\,0.23 & 41.68\,$\pm$\,0.43 & 76.06\,$\pm$\,0.13 & \underline{75.19\,$\pm$\,0.20} \\
& FedRAPP & \textbf{63.70\,$\pm$\,0.13} & \textbf{64.59\,$\pm$\,0.13} & 86.59\,$\pm$\,0.19 & \textbf{87.63\,$\pm$\,0.16} & \textbf{83.87\,$\pm$\,0.18} & \textbf{86.12\,$\pm$\,0.23} & \textbf{42.50\,$\pm$\,0.13} & \textbf{43.27\,$\pm$\,0.08} & \textbf{76.65\,$\pm$\,0.15} & \textbf{76.91\,$\pm$\,0.17} \\

\midrule

\multirow{9}{*}{ADAPROMPT}
& None-Fed & 62.97\,$\pm$\,0.41 & 64.72\,$\pm$\,0.41 & 87.47\,$\pm$\,0.54 & 86.90\,$\pm$\,0.01 & 84.54\,$\pm$\,0.23 & \underline{84.30\,$\pm$\,0.41} & 42.66\,$\pm$\,0.13 & 43.09\,$\pm$\,0.05 & 76.70\,$\pm$\,0.56 & 75.79\,$\pm$\,0.39 \\
& FedAvg & 63.17\,$\pm$\,0.03 & 62.75\,$\pm$\,0.02 & 86.60\,$\pm$\,0.60 & 86.27\,$\pm$\,0.20 & 84.33\,$\pm$\,0.19 & 81.09\,$\pm$\,0.08 & 42.75\,$\pm$\,0.10 & 42.56\,$\pm$\,0.35 & 76.59\,$\pm$\,0.51 & 75.36\,$\pm$\,0.24 \\
& FedAvgM & 63.30\,$\pm$\,0.67 & 62.25\,$\pm$\,0.11 & 87.37\,$\pm$\,0.21 & 85.77\,$\pm$\,0.69 & 84.31\,$\pm$\,0.05 & 81.75\,$\pm$\,0.19 & 42.96\,$\pm$\,0.03 & 42.04\,$\pm$\,0.20 & 76.85\,$\pm$\,0.37 & 74.81\,$\pm$\,0.05 \\
& FedProx & 63.44\,$\pm$\,0.32 & 62.41\,$\pm$\,0.36 & 87.21\,$\pm$\,0.08 & 85.61\,$\pm$\,0.48 & 84.40\,$\pm$\,0.40 & 81.96\,$\pm$\,0.03 & 43.07\,$\pm$\,0.13 & 40.82\,$\pm$\,0.16 & \textbf{76.90\,$\pm$\,0.29} & 74.70\,$\pm$\,0.24 \\
& FedAMP & \underline{63.68\,$\pm$\,0.17} & \underline{64.74\,$\pm$\,0.35} & \underline{87.86\,$\pm$\,0.19} & \underline{87.12\,$\pm$\,0.10} & 81.82\,$\pm$\,0.31 & 83.32\,$\pm$\,0.03 & \textbf{43.35\,$\pm$\,0.29} & \underline{43.15\,$\pm$\,0.39} & 76.14\,$\pm$\,0.16 & \underline{75.98\,$\pm$\,0.43} \\
& pFedGraph & \textbf{63.83\,$\pm$\,0.44} & 64.45\,$\pm$\,0.58 & 87.00\,$\pm$\,0.49 & 86.55\,$\pm$\,0.21 & 82.48\,$\pm$\,0.23 & 82.73\,$\pm$\,0.11 & 42.67\,$\pm$\,0.52 & 42.27\,$\pm$\,0.04 & 76.42\,$\pm$\,0.26 & 75.47\,$\pm$\,0.10 \\
& pFedMoAP & 60.35\,$\pm$\,0.61 & 61.24\,$\pm$\,0.71 & 83.63\,$\pm$\,0.54 & 84.80\,$\pm$\,0.47 & 80.59\,$\pm$\,0.38 & 80.63\,$\pm$\,0.64 & 39.36\,$\pm$\,0.23 & 40.49\,$\pm$\,0.12 & 74.09\,$\pm$\,0.76 & 73.81\,$\pm$\,0.64 \\
& FedCTTA & 63.67\,$\pm$\,0.61 & 64.16\,$\pm$\,0.12 & 86.99\,$\pm$\,0.09 & 86.88\,$\pm$\,0.66 & \underline{85.39\,$\pm$\,0.16} & 84.24\,$\pm$\,0.03 & 42.81\,$\pm$\,0.31 & 42.12\,$\pm$\,0.39 & 76.35\,$\pm$\,0.01 & 75.27\,$\pm$\,0.18 \\
& FedRAPP & 63.45\,$\pm$\,0.12 & \textbf{65.61\,$\pm$\,0.09} & \textbf{87.90\,$\pm$\,0.10} & \textbf{87.23\,$\pm$\,0.14} & \textbf{85.53\,$\pm$\,0.07} & \textbf{84.66\,$\pm$\,0.05} & \underline{43.18\,$\pm$\,0.11} & \textbf{43.27\,$\pm$\,0.12} & \underline{76.87\,$\pm$\,0.10} & \textbf{76.12\,$\pm$\,0.16} \\

\bottomrule

\end{tabular}
\end{adjustbox}
\end{table*}

\begin{table*}[t]
\centering
\caption{Performance with the CLIP ResNet-50 (RN50) image encoder on CLIP datasets under IID and non-IID partitions (Part 2). Mean top-1 accuracy\,$\pm$\,standard deviation over three random seeds. \textbf{Best} and \underline{second-best} per column within each block.}
\label{tab:rn50-part2}

\footnotesize
\setlength{\tabcolsep}{2.0pt}
\renewcommand{\arraystretch}{1.0}

\begin{adjustbox}{max width=\textwidth}
\begin{tabular}{@{}llcccccccc@{}}

\toprule

\multirow{2}{*}{TTA}
& \multirow{2}{*}{Fed.}
& \multicolumn{2}{c}{UCF101}
& \multicolumn{2}{c}{SUN397}
& \multicolumn{2}{c}{FGVCAircraft}
& \multicolumn{2}{c}{StanfordCars} \\

\cmidrule(l{1pt}r{1pt}){3-4}
\cmidrule(l{1pt}r{1pt}){5-6}
\cmidrule(l{1pt}r{1pt}){7-8}
\cmidrule(l{1pt}r{1pt}){9-10}

& & IID & Non-IID & IID & Non-IID & IID & Non-IID & IID & Non-IID \\

\midrule

\multirow{9}{*}{SWAPPROMPT}
& None-Fed & 58.99\,$\pm$\,0.44 & 58.80\,$\pm$\,0.59 & 59.23\,$\pm$\,0.23 & 59.37\,$\pm$\,0.09 & 17.84\,$\pm$\,0.45 & 17.96\,$\pm$\,0.39 & 57.44\,$\pm$\,0.13 & 58.01\,$\pm$\,0.22 \\
& FedAvg & 59.70\,$\pm$\,0.19 & 58.42\,$\pm$\,0.41 & 59.72\,$\pm$\,0.19 & 58.76\,$\pm$\,0.14 & 18.29\,$\pm$\,0.01 & 17.53\,$\pm$\,0.13 & 58.34\,$\pm$\,0.17 & 57.63\,$\pm$\,0.14 \\
& FedAvgM & \underline{60.56\,$\pm$\,0.27} & 58.11\,$\pm$\,0.36 & 60.12\,$\pm$\,0.57 & 58.08\,$\pm$\,0.05 & \underline{19.22\,$\pm$\,0.16} & 16.87\,$\pm$\,0.02 & \underline{59.23\,$\pm$\,0.32} & 57.25\,$\pm$\,0.15 \\
& FedProx & 59.98\,$\pm$\,0.59 & 57.41\,$\pm$\,0.12 & 59.76\,$\pm$\,0.16 & 57.42\,$\pm$\,0.53 & 18.64\,$\pm$\,0.09 & 16.14\,$\pm$\,0.68 & 58.48\,$\pm$\,0.19 & 56.73\,$\pm$\,0.33 \\
& FedAMP & 60.44\,$\pm$\,0.44 & \underline{59.27\,$\pm$\,0.35} & \underline{60.33\,$\pm$\,0.17} & \textbf{60.99\,$\pm$\,0.24} & 18.90\,$\pm$\,0.16 & 18.08\,$\pm$\,0.51 & 58.89\,$\pm$\,0.76 & \underline{59.41\,$\pm$\,0.51} \\
& pFedGraph & 59.71\,$\pm$\,0.37 & 59.08\,$\pm$\,0.14 & 60.13\,$\pm$\,0.59 & 60.73\,$\pm$\,0.20 & 18.72\,$\pm$\,0.17 & \underline{18.10\,$\pm$\,0.22} & 58.48\,$\pm$\,0.48 & 57.77\,$\pm$\,0.17 \\
& pFedMoAP & 57.06\,$\pm$\,0.89 & 56.31\,$\pm$\,0.01 & 56.52\,$\pm$\,0.61 & 57.75\,$\pm$\,0.35 & 15.75\,$\pm$\,0.16 & 14.47\,$\pm$\,0.72 & 55.66\,$\pm$\,0.37 & 55.99\,$\pm$\,0.90 \\
& FedCTTA & 59.80\,$\pm$\,0.57 & 58.23\,$\pm$\,0.14 & 59.90\,$\pm$\,0.57 & 59.40\,$\pm$\,0.17 & 18.58\,$\pm$\,0.20 & 17.22\,$\pm$\,0.30 & 58.38\,$\pm$\,0.36 & 58.80\,$\pm$\,0.17 \\
& FedRAPP & \textbf{60.67\,$\pm$\,0.12} & \textbf{59.72\,$\pm$\,0.16} & \textbf{60.42\,$\pm$\,0.16} & \underline{60.97\,$\pm$\,0.10} & \textbf{19.43\,$\pm$\,0.17} & \textbf{18.14\,$\pm$\,0.16} & \textbf{59.30\,$\pm$\,0.14} & \textbf{59.46\,$\pm$\,0.19} \\

\midrule

\multirow{9}{*}{ADAPROMPT}
& None-Fed & 59.97\,$\pm$\,0.52 & 60.71\,$\pm$\,0.02 & 59.42\,$\pm$\,0.22 & 61.26\,$\pm$\,0.32 & 18.53\,$\pm$\,0.61 & 18.30\,$\pm$\,0.31 & \underline{59.25\,$\pm$\,0.53} & 58.08\,$\pm$\,0.07 \\
& FedAvg & 60.05\,$\pm$\,0.70 & 59.04\,$\pm$\,0.04 & 60.09\,$\pm$\,0.67 & 58.91\,$\pm$\,0.07 & 18.92\,$\pm$\,0.32 & 17.68\,$\pm$\,0.52 & 59.07\,$\pm$\,0.20 & 58.54\,$\pm$\,0.62 \\
& FedAvgM & \underline{61.47\,$\pm$\,0.73} & 58.30\,$\pm$\,0.40 & 60.71\,$\pm$\,0.49 & 58.97\,$\pm$\,0.79 & 19.26\,$\pm$\,0.00 & 17.19\,$\pm$\,0.32 & \textbf{60.11\,$\pm$\,0.43} & 58.10\,$\pm$\,0.12 \\
& FedProx & 60.33\,$\pm$\,0.14 & 57.59\,$\pm$\,0.17 & 59.82\,$\pm$\,0.28 & 57.81\,$\pm$\,0.11 & 19.28\,$\pm$\,0.03 & 16.14\,$\pm$\,0.06 & 58.65\,$\pm$\,0.45 & 57.23\,$\pm$\,0.03 \\
& FedAMP & 61.35\,$\pm$\,0.55 & \underline{60.78\,$\pm$\,0.19} & \textbf{60.84\,$\pm$\,0.31} & \underline{61.79\,$\pm$\,0.25} & \underline{19.76\,$\pm$\,0.15} & \underline{18.73\,$\pm$\,0.13} & 59.08\,$\pm$\,0.41 & \underline{59.98\,$\pm$\,0.53} \\
& pFedGraph & 60.19\,$\pm$\,0.00 & 59.93\,$\pm$\,0.20 & 60.62\,$\pm$\,0.44 & 60.97\,$\pm$\,0.72 & 19.15\,$\pm$\,0.02 & 18.69\,$\pm$\,0.09 & 58.86\,$\pm$\,0.28 & 57.81\,$\pm$\,0.31 \\
& pFedMoAP & 57.78\,$\pm$\,0.54 & 56.58\,$\pm$\,0.08 & 56.98\,$\pm$\,0.73 & 58.30\,$\pm$\,0.29 & 16.38\,$\pm$\,0.66 & 15.33\,$\pm$\,0.09 & 56.54\,$\pm$\,0.32 & 56.96\,$\pm$\,0.09 \\
& FedCTTA & 60.46\,$\pm$\,0.42 & 58.70\,$\pm$\,0.37 & 60.74\,$\pm$\,0.07 & 59.61\,$\pm$\,0.29 & 19.21\,$\pm$\,0.29 & 17.27\,$\pm$\,0.25 & 58.40\,$\pm$\,0.60 & 59.22\,$\pm$\,0.28 \\
& FedRAPP & \textbf{61.55\,$\pm$\,0.07} & \textbf{60.99\,$\pm$\,0.15} & \underline{60.81\,$\pm$\,0.08} & \textbf{61.90\,$\pm$\,0.12} & \textbf{19.93\,$\pm$\,0.16} & \textbf{19.79\,$\pm$\,0.13} & 59.13\,$\pm$\,0.04 & \textbf{60.37\,$\pm$\,0.10} \\

\bottomrule

\end{tabular}
\end{adjustbox}
\end{table*}

\section{Per-Client Accuracy: Improving Weak Clients Without Harming Strong Ones}
\begin{figure*}[t]
    \centering
    \includegraphics[width=0.8\linewidth]{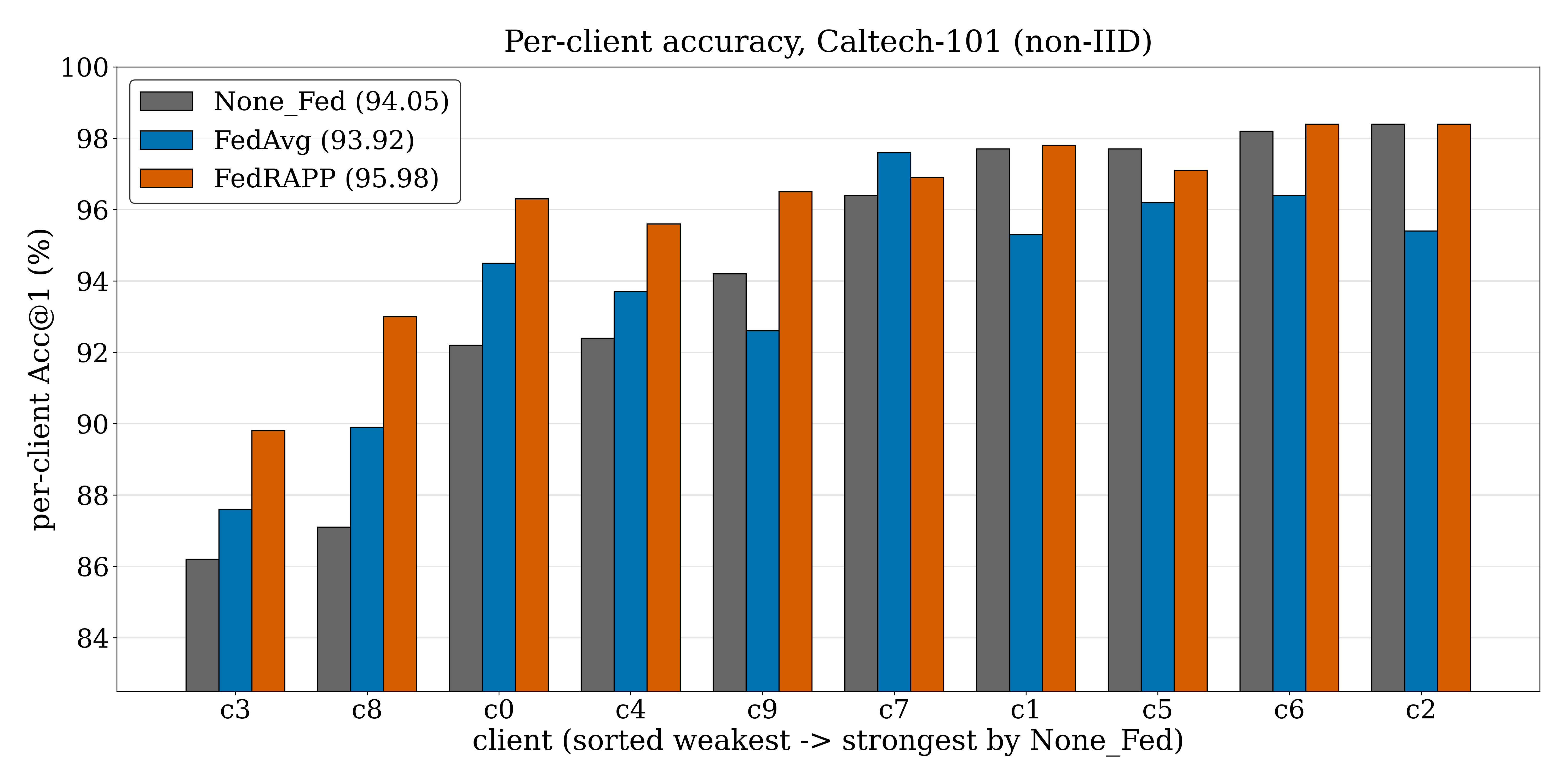}
    \caption{Per-client top-1 accuracy on Caltech-101 under non-IID (pathological)
partitioning, comparing None-Fed (local-only adaptation), FedAvg, and
FedRAPP. Clients are sorted from weakest to strongest by their None-Fed
accuracy. FedRAPP improves on the weak clients without degrading the strong ones,
whereas FedAvg's uniform aggregation drags several strong clients down.}
    \label{fig:pets-per-client-acc}
\end{figure*}

In Figure~\ref{fig:pets-per-client-acc} we analyze the effect of FedRAPP's performance at client-level on Caltech-101 dataset under non-iid setup. We compare the client-level performance of FedRAPP with FedAVG and under no aggregation (None-Fed). The results in Figure~\ref{fig:pets-per-client-acc} come from a single run with one seed, so the averages in its legend differ slightly from the mean values over three random seeds reported in Table~\ref{tab:clip-full-part1}. We observe that FedAvg improves the performance of weaker clients, however degrades the performance of the clients with stronger performance. We attribute this to the naive aggregation strategy of FedAvg as forcing every client toward a single shared prompt induces negative transfer. In contrast, FedRAPP improves the performance of the weaker clients while also keeping the performance of the stronger clients stable by personalizing collaboration per client and weighting it by reliability and drift similarity.

\section{Federated Aggregation Under DART}
\label{sec:dart-federation}
DART differs from the other test-time methods in this study by maintaining four distinct sets of prompts rather than one. It learns a separate learnable text prompt for every class, together with a single learnable visual prompt inserted into the image encoder, and adapts both on each incoming sample. Alongside these, it carries a second pair of knowledge-retention prompts, one per class on the text side and one per class on the visual side, which accumulate a running average of the adapted prompts and are blended back in before the next sample arrives, so that information from earlier test images is preserved. Only the two adapted prompt sets are trainable; the retention pair is a memory. For consistency with the other methods, all main-table results aggregate the two tunable sets and keep the retention memory on-device, local to the clients. Because this configuration proved unusually fragile, we additionally evaluate alternative choices of federation target on the OxfordPets dataset (Table~\ref{tab:dart-federation-target}). Each entry in Table~\ref{tab:dart-federation-target} comes from a single run with one seed, whereas Table~\ref{tab:clip-full-part1} reports the mean\,$\pm$\,standard deviation over three random seeds. For this reason, the first row of Table~\ref{tab:dart-federation-target}, which uses the main-table configuration, differs slightly from the DART entries for OxfordPets in Table~\ref{tab:clip-full-part1} (\eg, 48.91\% \vs 48.14\% for FedRAPP under IID). Withholding the class-specific text prompts and sharing only the visual prompt recovers a large part of the loss under both partitions, confirming that averaging class-specific text prompts across clients is the dominant failure mode. Next, we test a configuration where both the tunable text and visual prompts are aggregated as before, but the retention memory is re-initialized from the aggregated prompts in each round. This setting is harmful under plain averaging (FedAvg), as it repeatedly discards the client's own accumulated history, whereas the results improve for the personalized methods (FedAMP and FedRAPP). The strongest results come from aggregating the retention memory while keeping the tunable prompts local. In this case, the text retention prompts are aggregated class-wise, while the visual class-wise retention prompts are averaged (visual pooled) before being redistributed. This setup improves both the IID and non-IID performance under all aggregation methods, with the best results obtained by FedRAPP, which reaches 72.23\% and 83.85\% under the IID and non-IID partitions, respectively. However, these results remain below those of AdaPrompt and SwapPrompt (Table~\ref{tab:clip-full-part1}). We observe that not averaging the visual class-wise retention prompts degrades the results in five of the six settings, for example, from 62.42 to 56.88 with FedAvg under the IID partition. Finally, sharing only the text retention prompts does not obtain the best results for DART, except for FedAMP under the IID partition (71.00). DART's per-class, dual-modal prompt representation is therefore highly sensitive to which prompt set is shared, and no federation target we tested recovers standalone performance.

\begin{table}[t]
\centering
\caption{Analysis of federated techniques using DART as local adaptation on OxfordPets. Rows differ only in
\emph{which} of DART's four prompt sets is sent to the server (for aggregation); the local
adaptation procedure is identical throughout. The first federated row
(tunable text + tunable visual) is the configuration used across all the main tables.}
\label{tab:dart-federation-target}
\footnotesize
\setlength{\tabcolsep}{3.5pt}
\renewcommand{\arraystretch}{1.25}
\begin{adjustbox}{max width=\columnwidth}
\begin{tabular}{@{}lcccccc@{}}
\toprule
\multirow{2}{*}{Aggregated Prompts}
& \multicolumn{2}{c}{FedAvg} & \multicolumn{2}{c}{FedAMP} & \multicolumn{2}{c}{FedRAPP} \\
\cmidrule(l{2pt}r{2pt}){2-3} \cmidrule(l{2pt}r{2pt}){4-5} \cmidrule(l{2pt}r{2pt}){6-7}
& IID & Non-IID & IID & Non-IID & IID & Non-IID \\
\midrule
Tunable text + tunable visual      & 26.90 & 32.10 & 49.20 & 48.91 & 48.91 & 50.53 \\
Tunable visual only                & 48.57 & 60.48 & 59.06 & 76.34 & 59.64 & 77.23 \\
Tunable text + Tunable visual + Retention re-seeded  & \phantom{0}7.85 & \phantom{0}8.78 & 57.10 & 71.87 & 57.36 & 77.31 \\
Tunable visual only, retention re-seeded & 36.63 & 35.57 & 47.86 & 71.95 & 47.53 & 72.16 \\
Retention Prompts (visual pooled)     & 62.42 & 71.76 & 67.89 & 82.39 & \textbf{72.23} & \textbf{83.85} \\
Retention Prompts (visual class-wise) & 56.88 & 67.08 & 68.87 & 78.66 & 68.99 & 78.81 \\
Text retention Prompts                & 55.57 & 68.74 & 71.00 & 79.75 & 70.93 & 81.45 \\
\bottomrule
\end{tabular}
\end{adjustbox}
\end{table}

\section{Effect of the Attention Module in pFedMoAP}
\label{sec:pfedmoap-gating}

pFedMoAP~\cite{luo2025mixture} treats federated prompt learning as a mixture-of-experts problem:
each client holds its own prompt as a local expert, downloads a set of peer prompts as
non-local experts, and fuses them into an enhanced text feature used for classification.
In the original design, this fusion is performed by a lightweight single-head attention
module that is trained on each client and is never aggregated; only the prompts are
aggregated. The original work trains this attention module with labeled client data.
In our source-free test-time setting, no labels are available. Therefore, in all
pFedMoAP results of the main paper and the other supplementary tables, we replace the
trainable attention module with a \textbf{dot-product} variant, which scores each peer
prompt by its direct similarity to the local prompt and has no learnable parameters.
In Table~\ref{tab:pfedmoap-gating}, we compare this variant with the original
\textbf{trainable gate} setup on three datasets. In the trainable gate setup, the
attention module is randomly initialized on each client and co-trained with the prompt
on the same unlabeled test-time objective that the local adaptation method minimizes,
and the loss is computed on the fused logits.

\begin{table}[t]
\centering
\caption{Effect of pFedMoAP's trainable attention gate under label-free
test-time adaptation on three CLIP benchmarks. Top-1 accuracy with AdaPrompt as the
local adaptation method. \emph{Dot-product} is the variant used in the main tables and
scores peers by prompt similarity; \emph{trainable gate} is the original pFedMoAP setup,
with the attention module trained on the unlabeled test-time objective. Single seed;
therefore, the dot-product values differ slightly from Table~\ref{tab:clip-full-part1}.}
\label{tab:pfedmoap-gating}
\small
\setlength{\tabcolsep}{6pt}
\renewcommand{\arraystretch}{1.15}
\begin{tabular*}{\linewidth}{@{\extracolsep{\fill}}lcccc@{}}
\toprule
\multirow{2}{*}{Dataset}
& \multicolumn{2}{c}{Dot-product}
& \multicolumn{2}{c}{Trainable gate} \\
\cmidrule(l{2pt}r{2pt}){2-3} \cmidrule(l{2pt}r{2pt}){4-5}
& IID & Non-IID & IID & Non-IID \\
\midrule
OxfordPets & 84.40 & 85.16 & 60.23 & 70.46 \\
Caltech101 & 89.28 & 91.90 & 70.43 & 77.20 \\
Flower102  & 67.03 & 68.31 & 36.09 & 47.46 \\
\midrule
Mean       & 80.24 & 81.79 & 55.58 & 65.04 \\
\bottomrule
\end{tabular*}
\end{table}

The trainable gate reduces accuracy on every dataset and partition, by 14.70 to 30.94
points. On average, accuracy drops from 80.24 to 55.58 under the IID partition and from
81.79 to 65.04 under the non-IID partition. The reason is that a randomly initialized
attention module has no reliable signal to learn from in this setting. In the original
setting of pFedMoAP, the module is trained with labeled client data, which provides a
supervised gradient strong enough to train it. Under test-time adaptation, the only
available gradient comes from an unsupervised objective, such as entropy, on a small
unlabeled batch, which is too weak to train the module from scratch. As a result, the
untrained module adds noise to the fused logits throughout adaptation, and the prompt is
optimized on these noisy logits. For this reason, we use the dot-product variant in all
other experiments.

\begin{figure}[htbp]
  \centering
  \includegraphics[width=\linewidth, height=4cm]{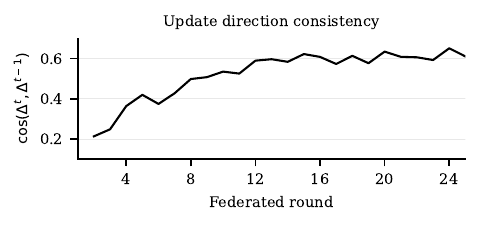}%
  \caption{Similarity between consecutive round prompt updates.}
  \label{cosine}
\end{figure}

\section{Performance analysis on ImageNet variants for TPT and DART}
\label{sec:imagenet-tpt-dart}

Table~\ref{tab:imagenet-variants} of the main paper reports the ImageNet-variant
results for SwapPrompt and AdaPrompt. Table~\ref{tab:imagenet-tpt-dart} provides the remaining TPT and DART results, reported as mean\,$\pm$\,standard deviation. The two blocks reproduce, under natural distribution shift, the behaviour already observed on the CLIP benchmarks. TPT is by far the weakest standalone
adaptation method, yet it
benefits from collaboration: every federated method improves on the
local-only baseline, and FedRAPP obtains the best average in the
block (29.98 and 29.79 under the IID and non-IID partitions, compared with 9.54 and 9.74 for None-Fed). DART shows the opposite pattern. Its local-only (None-Fed) configuration is the
strongest entry in the whole table and every federated variant
falls far below it under both partitions, because DART jointly tunes a visual prompt and class-specific text prompts, and aggregating them across clients disrupts this joint adaptation even when clients share the same classes; see Section~\ref{sec:dart-federation} and Table~\ref{tab:dart-federation-target} for the detailed
federation-target study for DART. Among the federated methods, FedRAPP is nonetheless the
best under the IID partition and the second-best under the non-IID partition, confirming that
reliability-weighted personalized aggregation is among the least destructive
aggregation strategies even when the underlying prompt representation resists
sharing.

\section{Stability of Prompt updates}
In Figure \ref{cosine}, we show the similarity of the consecutive prompts updates for the OxfordPets dataset across 10 clients using FedRAPP for non-iid scenario. We calculate the cosine similarity of the prompt updates per client for consecutive rounds and show the average of the cosine similarities for all the clients per round. We observe that prompt updates become increasingly consistent (more similar in their direction) and stable across the federated rounds, as shown by the rising cosine similarity.

\end{document}